%% file: main.tex
\documentclass[10pt,twocolumn,letterpaper]{article}

\usepackage[pagenumbers]{cvpr} 

\input{preamble}

\definecolor{cvprblue}{rgb}{0.21,0.49,0.74}
\usepackage[pagebackref,breaklinks,colorlinks,allcolors=cvprblue]{hyperref}

\def\paperID{8493} 
\def\confName{CVPR}
\def\confYear{2025}

\title{
HierarQ: Task-Aware Hierarchical Q-Former for Enhanced Video Understanding
}

\author{Shehreen Azad\textsuperscript{1}  \qquad  \qquad Vibhav Vineet\textsuperscript{2} \vspace{3pt} \qquad \qquad Yogesh Singh Rawat\textsuperscript{1} \\
\textsuperscript{1}Center for Research in Computer Vision, University of Central Florida; \qquad \textsuperscript{2}Microsoft Research\\
\href{https://sacrcv.github.io/HierarQ-website/}{Project page}}

\begin{document}

\twocolumn[{%
\renewcommand\twocolumn[1][]{#1}%
\maketitle
\vspace{-8mm}
\begin{center}
    \centering
    \captionsetup{type=figure}
    \vspace{-10pt}
    \includegraphics[width=\linewidth]{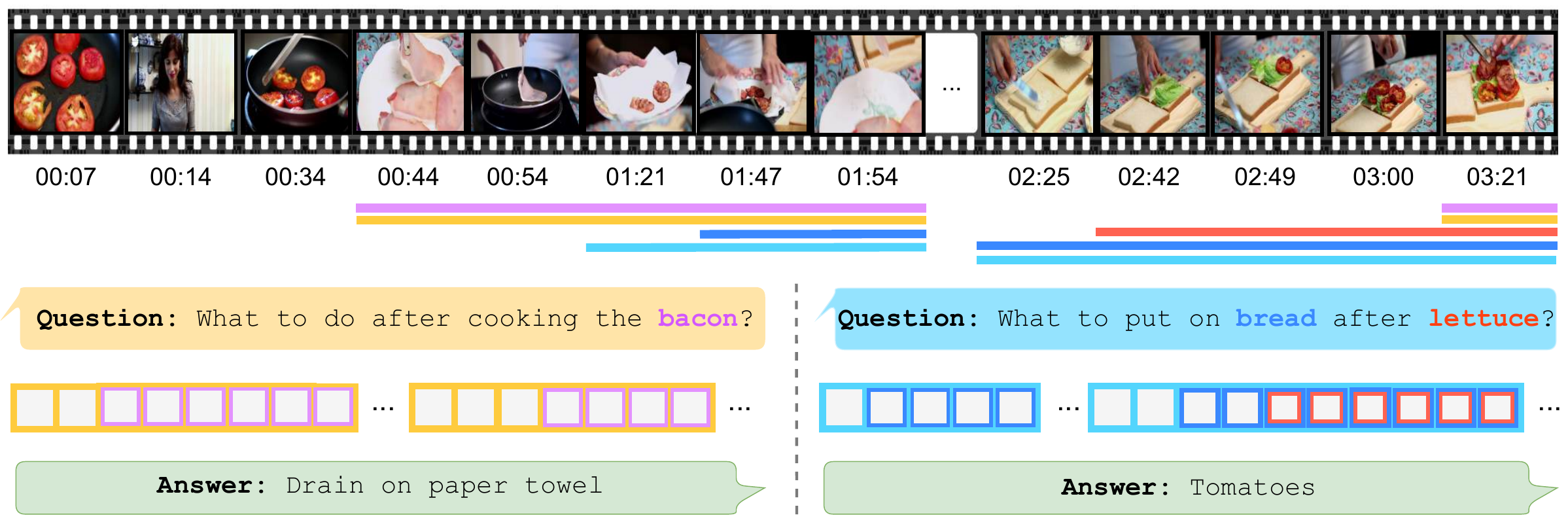}
    \captionof{figure}{\textbf{Effectiveness of HierarQ in capturing task-relevant information.} HierarQ adaptively focuses on task-relevant video segments, achieving a task-aware, comprehensive understanding. Here, color-coded frames are shown to demonstrate how entity-focused information complements the broader prompt-relevant context, enhancing overall video relevance and understanding.
    }
    \label{fig:teaser}
\end{center}%
}]

\input{sec/0_abstract}
\input{sec/1_intro}

\input{sec/2_rel_works}

\input{sec/3_method}

\input{sec/4_experiments}
\input{sec/5_analysis}

\input{sec/6_conclusion}

{
    \small
    \bibliographystyle{ieeenat_fullname}
    \bibliography{main}
}

\input{sec/X_suppl}

\end{document}

%% file: preamble.tex
\usepackage{multirow}
\usepackage{tabularx}
\usepackage{tablefootnote}
\usepackage{colortbl}
\usepackage{amsmath}
\usepackage{footnote}
\makesavenoteenv{figure}

\usepackage{textcomp}
 \newcommand{\mytexttilde}{\raisebox{0.5ex}{\texttildelow}}

%% file: sec/0_abstract.tex
\begin{abstract}
Despite advancements in multimodal large language models (MLLMs), current approaches struggle in medium-to-long video understanding due to frame and context length limitations. As a result, these models often depend on  frame sampling, which risks missing key information over time and lacks task-specific relevance.
To address these challenges, we introduce \textbf{HierarQ}, a task-aware hierarchical Q-Former based framework that sequentially processes frames to bypass the need for frame sampling, 
while avoiding LLM's context length limitations. We introduce a lightweight two-stream language-guided feature modulator to incorporate task awareness in video understanding, with the entity stream capturing frame-level object information within a short context and the scene stream identifying their broader interactions over longer period of time. 
Each stream is supported by dedicated memory banks which enables our proposed \textbf{Hierar}chical \textbf{Q}uerying transformer (HierarQ) to effectively capture 
short and long-term context. Extensive evaluations on \textbf{10} video benchmarks across video understanding, question answering, and captioning tasks demonstrate HierarQ’s state-of-the-art performance across most datasets, proving its robustness and efficiency for comprehensive video analysis.

\end{abstract}
\vspace{-10pt}

%% file: sec/1_intro.tex
\vspace{-7pt}
\section{Introduction}
\label{sec:sunday_intro}
The rapid progress in Large Language Models (LLMs) has significantly boosted AI capabilities, particularly in text generation and complex reasoning \cite{brown2020language, wei2021finetuned, touvron2023llama, touvron2023llama2, radford2018improving, radford2019language, chowdhery2023palm, devlin2018bert, raffel2020exploring}. Expanding these models into multimodal applications, Multimodal LLMs (MLLMs) have shown strong performance across image and short video tasks like captioning, question answering, and segmentation \cite{li2023blip, dai2023instructblip, li2023videochat, li2025videomamba, sun2019videobert, chen2023videollm, wang2023visionllm, zhu2023minigpt4, chen2023minigptv2, schiappa2024probing, schiappa2024robustness, azad2025understandingdepthheightperception}. 
However, as video content lengthens, these models encounter significant challenges due to context length limitations, which restrict their ability to process multiple frames and capture complex, extended temporal interactions.  
Although LLMs with extended context lengths are emerging \cite{dubey2024llama, abdin2024phi, team2024gemini}, they are computationally intensive and often fail to meet their theoretical context length promises \cite{an2023leval, hsieh2024ruler, li2024longcontext}.

To overcome context-length bottleneck, strategies like frame sampling \cite{yu2024self, lin2023videollava, zhang2023videollama} and spatio-temporal pooling \cite{maaz-etal-2024-video, luo2023valley} help reduce token loads but risk losing important temporal information, especially in longer videos. Methods using concatenated frame embeddings \cite{li2023blip, dai2023instructblip} and token compression \cite{song2024moviechat, song2024moviechat+, he2024ma, ren2024timechat, ren2023testa} often oversimplify complex sequences due to such coarse compression techniques, potentially losing crucial details. Timestamp reliance \cite{ren2024timechat} limits response generation without specific time markers, while video segmentation \cite{cheng2024enhancing, qian2024streaming, ge2022bridging} can disrupt narrative continuity. 
Additionally, these approaches lack task-level relevance, causing the models to process all frames blindly, leading to less effective information prioritization.

To address the aforementioned limitations, we introduce \textbf{HierarQ}, a task-aware \textbf{Hierar}chical \textbf{Q}uerying transformer based framework that processes videos auto-regressively without frame sampling. This setup preserves efficient processing while allowing HierarQ to maintain a task-focused, human-like cognitive approach, dynamically emphasizing relevant details based on task requirements (illustrated in Figure \ref{fig:teaser}) .

The core of our framework's task awareness lies in its two-stream text-guided feature modulator. While the entity-guided stream focuses on frame-level object understanding within a short context (entity-stream), the prompt-guided stream handles broader scene understanding across a longer context (scene-stream), capturing interactions between entities over time. By learning entity and scene information independently, our method allows the frame-level object information to complement the scene-level context, strengthening overall comprehension. 

To enhance each stream’s temporal modeling capabilities, we introduce dedicated memory banks. The entity stream is equipped with a short-term memory for retaining frame-specific object information; whereas the scene-stream has a long-term memory to capture the broader context of object interactions. These memory banks work together to support a balanced understanding of both immediate details and broader context, guided by our \textbf{Hierar}chical \textbf{Q}uerying Transformer (\textbf{HierarQ}). HierarQ integrates entity-level insights into scene-level comprehension, achieving a nuanced understanding that effectively balances task relevance with temporal coherence.

We extensively evaluate our method on multiple video tasks, covering Medium to Long Video Understanding with 3 benchmarks, Video Question Answering with 4 benchmarks, and Video Captioning with 3 benchmarks. HierarQ
achieves state-of-the-art results across most of the benchmarks and competitive performance on others, demonstrating its robustness and adaptability in diverse video understanding tasks.

Our contributions are as follows:
\begin{itemize}
    \item We introduce \textbf{HierarQ}, a task-aware Hierarchical Querying transformer based framework with short and long-term memory banks for enhanced video understanding.
    \item To introduce task-awareness we propose a lightweight two-stream feature modulator to dynamically modulate task-relevant frames, optimizing task-aware processing.
    \item Our method achieves state-of-the-art performance 
    across most of the \textbf{10} popular benchmarked datasets, while having competitive performance on others, demonstrating robustness in complex video comprehension.
\end{itemize}

%% file: sec/2_rel_works.tex
\vspace{-6pt}
\section{Related Works}
\label{sec:rel_works}
\vspace{-4pt}
\textbf{Multi-modal Large Language Models.} CLIP \cite{radford2021learningtransferablevisualmodels} introduced contrastive learning for image-text alignment, followed by several other works \cite{li2024llava, li2023blip, dai2023instructblip, zhu2023minigpt4, alayrac2022flamingo, awadalla2023openflamingo} to further enhance image-text understanding by bringing visual features closer to language space. Among these, BLIP-2’s Q-Former \cite{li2023blip} is a lightweight approach that effectively bridges visual and language modalities but, like other image-focused models, it struggles with the spatio-temporal complexities of videos. To address temporal dynamics, several video-based models \cite{li2023videochat, lin2023videollava, zhang2023videollama, maaz-etal-2024-video, xu2024pllavaparameterfreellava, li2025videomamba} have been proposed, but they remain constrained by the frame limits, making them more suitable for short videos.

\noindent\textbf{Medium to Long Video Understanding Models.} Medium to long term video understanding 
aims to capture extended patterns in videos typically exceeding $30$ seconds. Prior works in this area includes unimodal \cite{wu2021towards, patrick2021keeping, islam2022long, islam2023efficient, wu2022memvit, bolya2022token} and multimodal models \cite{song2024moviechat, song2024moviechat+, he2024ma, cheng2024enhancing, faure2024bridging, yu2024self, qian2024streaming, jin2023chatunivi, xu2024pllavaparameterfreellava}. Strategies like sparse sampling, temporal pooling, and token compression are commonly used to reduce redundancy, while also addressing LLM context length limits. Recently, memory-based methods have also gained attention \cite{song2024moviechat, he2024ma, cheng2024enhancing, faure2024bridging, balazevic2024memory, weng2024longvlm, zhang2024flash} that efficiently compresses historical information via parametric \cite{wu2022memvit} or non-parametric methods \cite{song2024moviechat, he2024ma}. Our approach extends these designs \cite{wu2022memvit, he2024ma, song2024moviechat} by integrating both short and long-term memory banks with large MLLMs, enhancing temporal modeling for effective video comprehension.

\noindent
\textbf{Task-aware Frame Processing.} To achieve task-aware processing, several models \cite{buch2022revisiting, lu2022lgdn, gao2023mist, yu2024self, Wang2024, ranasinghe2024understanding} use language-guided key-frame selection, effective in short videos but less so for longer videos where extended temporal relationships are essential. Key details may span multiple frames, and critical information could be missed if sampling is too sparse. Moreover, it is difficult to capture the full context with only some  frames, as essential details may also be contained within frames deemed less relevant. Inspired by \cite{narasimhan2021clip}, our approach addresses this by processing every frame, focusing more on task-relevant frames without excluding others, ensuring both comprehensive coverage and prioritized, task-specific information.

%% file: sec/3_method.tex
\vspace{-4pt}
\section{Method}
\label{sec:method}
\vspace{-4pt}

\begin{figure}
    \centering
    \includegraphics[width=.9\linewidth]{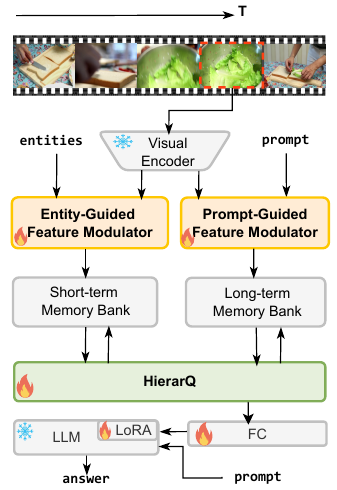}
    \caption{\textbf{Overview of our framework} that sequentially processes video frames, modulating task-relevant entity and scene features with a two-stream feature modulator. The proposed \textbf{HierarQ} (\textbf{Hierar}chical \textbf{Q}-Former)  with dedicated memory banks integrates these features, producing a refined understanding that is passed to an LLM for the final response. The flame and snowflake icons respectively denote trainable and frozen parameters.}
    \label{fig:main_fig}
    \vspace{-15pt}
\end{figure}

To enable effective task-aware video understanding without hitting LLM context length limits or input frame restrictions, we propose \textbf{HierarQ}, a task-aware \textbf{Hierar}chical \textbf{Q}uerying transformer based framework. Our model processes videos auto-regressively, and uses a language-guided two-stream feature modulator to enhance task relevance (Section \ref{subsec:vis}). Each stream is supported by dedicated memory banks, enabling our proposed hierarchical querying transformer (\textbf{HierarQ}) to model long-term temporal relationships effectively (Section \ref{subsec:qf}). Finally, the output from HierarQ is sent to an LLM to generate the response text (Section \ref{subsec:td}). Given an input video $V$ and prompt $T_P$, our framework performs reasoning to produce the text output $T_O$. The overview of our approach is illustrated in Figure \ref{fig:main_fig}.

\vspace{-2pt}
\subsection{Task-aware Visual Feature Extraction}
\label{subsec:vis}
Our model processes frames sequentially in an auto-regressive manner, enabling efficient comprehension of lengthy videos without needing simultaneous multi-frame processing.
Given a sequence of $T$ frames,  we pass each frame $v_i$ to a pre-trained frozen visual encoder $\mathcal{V}(\cdot)$ to get the visual features $F$ as Equation \ref{eq:vis_extr}.
\begin{equation}
    F = \{f_i = \mathcal{V} (v_i) | \forall i = 1, ..., T\} , f_i \in \mathbb{R}^{N \times D}.
    \label{eq:vis_extr}
    \vspace{-2pt}
\end{equation} 

\noindent Here, each visual feature $f_i$ has temporal ordering information associated with it through a position embedding layer. The raw features are then fed to a dual-stream language-guided feature modulator. The primary objective of this lightweight transformer-based modulator is to modulate or put more focus on the visual features that are more relevant to the given prompt. To achieve this, we introduce a two-stream language-guided feature modulator module containing an entity-level modulator $L_f^e$  (Section \ref{subsubsec:lfe}) to put more focus on frames containing individual objects and persons (\textit{e.g. entities}), and a scene-level modulator $L_f^s$ (Section \ref{subsubsec:sfe}) for capturing interactions between these entities in alignment with the text prompt.

\vspace{-6pt}
\subsubsection{Entity-guided Feature Modulator} 
\label{subsubsec:lfe}
\vspace{-4pt}
Given a text prompt $T_P$, we extract nouns that represent entities (specifically \textit{person} or \textit{object}) and then use BERT \cite{devlin2018bert} to get their text embeddings, denoting it as $T_P^e$. Entity-guided feature modulator ($L_f^e$) puts higher focus on frame features that contain those entities by applying cross-attention between the frame’s raw visual features $f_i$ and entity embeddings $T_P^e$. Here, the key and value come from $f_i$ and the query comes from $T_P^e$ and then we do cross attention (\textit{C. Attn}) as Equation \ref{eq:qkv_lfe}.
\begin{equation}
    f_i^e = C.Attn (Q, K, V) \simeq C.Attn (T_P^e,f_i,f_i).
    \label{eq:qkv_lfe}
    \vspace{-2pt}
\end{equation}

Through this cross-attention mechanism, $L_f^e$ puts more focus on relevant entities in the frame, producing the modulated entity-level frame feature $f_i^e \in \mathbb{R}^{N \times D}$, which is stored in a short-term memory bank $M_e$ 
for short-term contextual reference which is discussed later.

\vspace{-6pt}
\subsubsection{Prompt-guided Feature Modulator} 
\label{subsubsec:sfe}
\vspace{-4pt}
Given a text prompt $T_P$, we use BERT \cite{devlin2018bert} to get the text embeddings, denoting it as scene embedding $T_P^s$. Prompt-guided Feature Modulator ($L_f^s$) leverages $T_P^s$ to attend to the raw frame embedding $f_i$ that are more relevant to the given prompt. Unlike $L_f^e$, which focuses on 
objects and persons within individual frames, $L_f^s$ captures more nuanced, scene-level relationships between those entities (whether person
or objects) in each frame.
This allows $L_f^s$ to establish broader contextual insights that evolve over time across the video, making it especially useful for tasks that require understanding the dynamic progression of a scene.
Similar to $L_f^e$, in $L_f^s$, the key and value comes from $f_i$, whereas the query comes from $T_P^s$ and then we do cross attention as Equation \ref{eq:qkv_lfs}. 
\begin{equation}
    f_i^s = C.Attn (Q, K, V) \simeq C.Attn (T_P^s,f_i,f_i).
    \label{eq:qkv_lfs}
    \vspace{-4pt}
\end{equation}

\noindent
The resulting modulated scene-level feature $f_i^s \in \mathbb{R}^{N \times D}$, captures the relevant context from the frame that aligns to the prompt and is then stored in a long-term memory bank $M_s$ 
which retains these broader, evolving context across frames, providing a richer understanding of the video content in relation to the given text prompt. 

\subsection{HierarQ}
\label{subsec:qf}
\begin{figure}
    \centering
    \includegraphics[width=\linewidth]{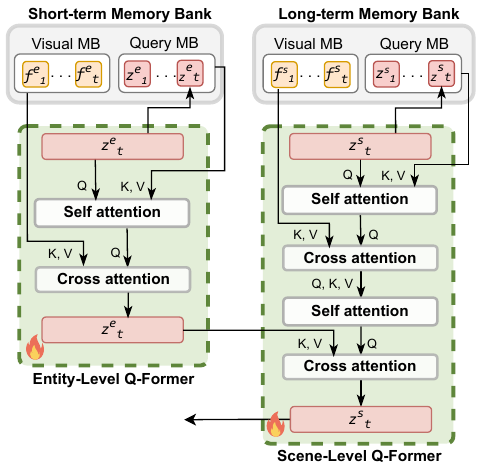}
    \caption{\textbf{Overview of HierarQ (Hierarchical Querying transformer).} It models the hierarchical relationship between then Entity-level Q-Former and Scene-level Q-Former, using dedicated memory banks to integrate short-term details with long-term context for enhanced video understanding.}
    \label{fig:hiqf}
    \vspace{-10pt}
\end{figure}

To capture evolving video dynamics, we introduce the \textbf{Hierar}chical \textbf{Q}uerying transformer (\textbf{HierarQ}), shown in Figure \ref{fig:hiqf}. Adapted from the original Q-Former \cite{li2023blip, dai2023instructblip}, HierarQ connects short-term frame-level entity details with broader scene-level context. By hierarchically integrating entity and scene features, HierarQ emulates human cognitive processing, where focused details contribute to a broader understanding. This design progressively enriches scene comprehension with entity-specific insights, enabling nuanced and temporally aware video understanding that adapts seamlessly across the video’s timeline.

HierarQ utilizes learnable queries $ z \in \mathbb{R}^{N \times D} $, where $N$ is the number of queries and $D$ is their dimension, outputting $32$ tokens per frame. It incorporates two distinct Q-formers: entity-level ($QF_e$) (Section \ref{subsubsec:eqf}) and scene-level ($QF_s$) (Section \ref{subsubsec:sqf}), each supported by dedicated memory banks to enhance hierarchical and temporal modeling. Rather than relying solely on the current frame’s embedding, HierarQ integrates these memory banks (inspired by \cite{he2024ma}) to enrich temporal context, enabling nuanced, long-term video comprehension.

\vspace{-6pt}
\subsubsection{Entity-level Q-Former}
\label{subsubsec:eqf}

Entity-level Q-Former ($QF_e$) is a standard Q-Former containing two attention submodules: (1) a self-attention layer for modeling interactions within the input queries, and (2) a cross-attention layer for interactions with modulated visual embeddings.  $QF_e$ is supported by a short-term memory bank $M_e$, containing both visual and query memories for effective short-range context.

\noindent
\textbf{Short-term memory bank} $\mathbf{(M_e)}$. The visual memory bank contains modulated entity-level visual features denoted as $F_t^e =$ \texttt{Concat} $[f_1^e, f_2^e, ..., f_t^e]$, $F_t^e \in \mathbb{R}^{tN \times D}$. Since, all cross-attention layer in $QF_e$ attends to the same visual feature, there is only one static visual memory bank. Given the input $z_t^e$ as query, the visual memory bank $F_t^e$ acts as key and value as Equation \ref{eq:qkv_e} and then we do cross-attention.

\begin{equation}
    Q = z_t^eW_q, \ K = F_t^eW_k, \ V = F_t^eW_v .
    \label{eq:qkv_e}
    \vspace{-3pt}
\end{equation}

\noindent The query memory bank contains input queries of each timestep denoted as $Z_t^e =$ \texttt{Concat} $[z_1^e, z_2^e, ..., z_t^e]$, $Z_t^e \in \mathbb{R}^{tN \times D}$. Unlike the static visual memory bank, $Z_t^e$ is unique to each layer of $QF_e$, each containing all the input queries to that layer upto the current timestep $t$. This allows $QF_e$ to look at past information about the video at increasing level of abstraction within a short context. Given the input $z_t^e$ as query, and the query memory bank $Z_t^e$ acts as key and value as Equation \ref{eq:qkv_eq} and then we do self-attention.

\begin{equation}
    Q = z_t^eW_q, \ K = Z_t^eW_k, \ V = Z_t^eW_v .
    \label{eq:qkv_eq}
    \vspace{-16pt}
\end{equation}

\subsubsection{Scene-level Q-Former}
\label{subsubsec:sqf}
\vspace{-4pt}
The Scene-level Q-Former ($QF_s$) extends $QF_e$ by adding two more submodules: (3) an additional self-attention for interactions between so far learned queries, and (4) cross-attention with queries from $QF_e$. This design allows $QF_s$ to leverage the long-term memory bank $M_s$ for broader context, integrating entity-level insights from $QF_e$. This layered approach enhances long-term video comprehension by combining broader scene-level context with detailed, complementary short-term entity information. Through this integration, entity-level insights enrich the scene-level understanding, allowing the model to achieve a more nuanced grasp of both immediate and extended temporal relationships in the video.

\noindent
\textbf{Long-term memory bank} ($\mathbf{M_s}$). The visual memory bank contains modulated scene-level visual features denoted as $F_t^s =$ \texttt{Concat} $[f_1^s, f_2^s, ..., f_t^s]$, $F_t^s \in \mathbb{R}^{tN \times D}$. Given the input $z_t^s$ of $QF_s$ as query, the visual memory bank $F_t^s$ acts as key and value as Equation \ref{eq:qkv_s} and then we do cross attention.
\begin{equation}
    Q = z_t^sW_q, \ K = F_t^sW_k, \ V = F_t^sW_v .
    \label{eq:qkv_s}
    \vspace{-3pt}
\end{equation}
  
\noindent The query memory bank contains input queries of each timestep denoted as $Z_t^s =$ \texttt{Concat} $[z_1^s, z_2^s, ..., z_t^s]$, $Z_t^s \in \mathbb{R}^{tN \times D}$. Same as $QF_e$, $QF_s$ also gets a unique query memory bank for each layer. Given input $z_t^s$ as query, the query memory bank $Z_t^s$ acts as key and value as Equation \ref{eq:qkv_sq} and then we do self-attention. 
\begin{equation}
    Q = z_t^sW_q, \ K = Z_t^sW_k, \ V = Z_t^sW_v .
    \label{eq:qkv_sq}
    \vspace{-3pt}
\end{equation}

\noindent The output from the cross-attention layer is the intermediate learned queries ($\hat{z}_t^s$) which then attends to itself through self-attention. Finally $\hat{z}_t^s$ interacts with learned query $z_t^e$ of $QF_e$ as Equation \ref{eq:qkv_sqf} and then we do cross-attention to get the final learned query $z_t^s$ as output of HierarQ. 
\begin{equation}
    Q = \hat{z}_t^sW_q, \ K = z_t^eW_k, \ V = z_t^eW_v .
    \label{eq:qkv_sqf}
    \vspace{-4pt}
\end{equation}
This hierarchical design allows refined interaction between entity and scene details over time, supporting a comprehensive and temporally aware understanding of video content.

\vspace{-6pt}
\subsubsection{Memory Bank Update Strategy} 
\vspace{-4pt}
To manage GPU memory and computational cost, which grow linearly with added frames, our model uses memory compression techniques tailored to the needs of entity- and scene-level information. For the short-term memory bank $M_e$ (both visual and query), we employ a First-in-First-Out (FIFO) queue similar to \cite{wu2022memvit, song2024moviechat}. This setup discards older entries when $M_e$ reaches capacity $M$, providing sufficient short-term context for frame-level entity details.

Conversely, the long-term memory bank $M_s$ requires long-term context to capture scene interactions over time. Here, a simple FIFO approach would lose critical scene continuity. Inspired by prior works \cite{bolya2022token, ren2023testa, jin2023chatunivi, song2024moviechat, he2024ma} demonstrating that merging similar tokens improves efficiency, we apply Memory Bank Compression (MBC) \cite{he2024ma} to compress $M_s$. MBC averages adjacent tokens ($f_t, f_{t+1}$) with high similarity (indexed by $k$), reducing redundancy while preserving temporal order
as shown in Equation \ref{eq:mbc}.
\begin{equation}
    k = \arg\max_t (cos(f_t, f_{t+1}), t \in [1,M]).
    \label{eq:mbc}
    \vspace{-4pt}
\end{equation}
By using FIFO for $M_e$ and MBC for $M_s$, our model effectively balances memory, preserving detailed short-term data for entities while retaining essential long-term context for scenes.

\subsection{Text decoding}
\label{subsec:td}
HierarQ’s final timestep output, which contains the entire historical context due to auto-regressive processing and memory banks, is projected through a fully connected layer 
to align with the LLM  dimension. This projection, combined with the prompt $T_P$, is fed to the LLM to generate the final output $T_O$. This setup reduces input text tokens to LLM from $N \times T$ to $N$, essentially addressing LLM context length limits and reducing memory demands. 

During training, we optimize trainable parameters  with video-text data using the standard cross entropy loss. Additional architectural details and training strategies have been provided in the supplementary material.

%% file: sec/4_experiments.tex
\section{Experiments}
\label{sec:experiments}

\subsection{Tasks, Datasets and Evaluation Metrics}
We evaluate our model on several tasks including, medium to long video understanding, short and long video question answering and video captioning following the same evaluation protocol as relevant prior works. 

\begin{table*}[t]
\centering
\small
\begin{minipage}{.54\textwidth}
    \caption{\textbf{Performance comparison of medium to long video understanding} on LVU dataset. The top-1 accuracy is reported. $\ddag$ indicates without LLM finetuning. \textbf{Best} and \underline{second-best} performances are highlighted.}
    \resizebox{0.988\linewidth}{!}{
    \begin{tabular}{l|ccc|c}
    \hline
    Model & Relation $\uparrow$ & Speak $\uparrow$ & Scene $\uparrow$ & Avg \\
    \hline
    VideoBERT   \cite{sun2019videobert}           & $52.8$      & $37.9$   & $54.9$   &   $48.5$  \\
    Obj\_T4mer\cite{wu2021towards}            & $54.8$      & $33.2$   & $52.9$  &  $47.0$                                            \\

    Orthoformer   \cite{patrick2021keeping}         & $50.0$      & $38.3$   & $66.3$   &    $51.5$                                          \\

    VIS4mer   \cite{islam2022long}             & $57.1$      & $40.8$   & $67.4$   &                                       $55.1$      \\
    LF-VILA \cite{sun2022long}& $61.5$&$41.3$&$68.0$&$56.9$\\
    TranS4mer \cite{islam2023efficient} & $59.5$ & $39.2$ & $70.9$ & $56.5$\\
    S5 \cite{wang2023selective}                    & $67.1$      & $42.1$   & $73.5$   &                                   $60.9$          \\
    Movies2Scene  \cite{chen2023movies2scenes} & $\mathbf{71.2}$&$42.2$&$68.2$&$60.5$\\
    VideoMamba \cite{li2025videomamba} & $62.5$ & $40.4$ & $70.4$& $57.8$\\
    MA-LMM  \cite{he2024ma}               & $58.2$      & $44.8$   & $80.3$   &$61.1$                                              \\
    
    \hline
    \rowcolor[gray]{0.92}
    \textbf{HierarQ $\mathbf{\ddag}$}                   &     $67.9$      & $\underline{48.7}$       & $\underline{83.8}$       & $\underline{66.8}$   \\ 
    \rowcolor[gray]{0.92}
    \textbf{HierarQ}                   &     \underline{$69.4$}      & $\mathbf{49.3}$       & $\mathbf{85.1}$       & $\mathbf{67.9}$    \\ 
    \hline
    \end{tabular}
    }
    \label{tab:lvu}
\end{minipage}
\hspace{6mm} 
\begin{minipage}{.37\textwidth}
    \caption{\textbf{Performance comparison of medium to long video understanding} on Breakfast and COIN datasets. The top-1 accuracy is reported.}
    \resizebox{.9\linewidth}{!}{
    \renewcommand{\arraystretch}{1.07}
    \begin{tabular}{l|c|c}
    \hline
        Model & Breakfast & COIN\\
        \hline
        Timeception \cite{hussein2019timeception}&$71.3$&-\\
        VideoGraph \cite{hussein2019videograph}& $69.5$&-\\
        GHRM \cite{zhou2021graph}& $75.5$&-\\
        Dist-Sprv \cite{lin2022learning} & $89.9$ & $90.0$\\
        ViS4mer \cite{islam2022long} & $88.2$& $88.4$\\
        TranS4mer \cite{islam2023efficient} & $90.3$&$89.2$\\
        S5 \cite{wang2023selective} & $90.7$&$90.8$\\
        FACT \cite{lu2024fact} & $84.5$ & -\\
        VideoMamba \cite{li2025videomamba} & ${94.3}$ & $86.2$\\
        MA-LMM \cite{he2024ma} & $93.0$&${93.2}$\\

        \hline
        \rowcolor[gray]{0.92}
        \textbf{HierarQ$\mathbf{\ddag}$} &$\underline{96.1}$&$\underline{94.6}$\\
        \rowcolor[gray]{0.92}
        \textbf{HierarQ} &$\mathbf{97.4}$&$\mathbf{96.0}$\\
        \hline
    \end{tabular}
    }
    \label{tab:breakfast}
\end{minipage}
\end{table*}

\begin{table*}[t!]
    \centering
    \small
    \begin{minipage}{.5\textwidth}
     \caption{\textbf{Performance comparison of long video question answering} on MovieChat-1k in Global (G) and Breakpoint (B) mode. The Accuracy (Acc.) and Score (Sc.) for GPT 3.5 assisted evaluation is reported. 
    $\dag$ indicates results reproduced in our environment.
    }
    \renewcommand{\arraystretch}{1.15}
    \resizebox{\linewidth}{!}{
    \begin{tabular}{l|cc|cc}
        \hline
         Model & G. Acc. & G. Sc. & B. Acc. & B. Sc.\\
         \hline
         Video Chat \cite{li2023videochat} & $57.8$& $3.0$& $46.1$& $2.3$\\
         Video LLaMA \cite{zhang-etal-2023-video} &  $51.7$& $2.7$& $39.1$& $2.0$\\
         Video-ChatGPT \cite{maaz-etal-2024-video} & $47.6$& $2.6$& $48.0$& $2.5$\\
         MovieChat \cite{song2024moviechat} & $62.3$ & $3.2$&$48.3$&$2.6$\\ 
         FVS+S3 \cite{wang2024hierarchical} & {$84.0$}& \underline{$4.6$}& {$73.5$}& {$4.0$}\\
         MA-LMM $\dag$ \cite{he2024ma} & $61.4$ & $3.2$ & $50.4$ & $2.7$\\
         \hline
         \rowcolor[gray]{0.92}
         \textbf{HierarQ$\mathbf{\ddag}$}  & $\underline{86.9}$ & $\mathbf{4.7}$ & $\underline{74.2}$ & $\underline{4.1}$\\
         \rowcolor[gray]{0.92}
         \textbf{HierarQ}  & $\mathbf{87.5}$ & $\mathbf{4.7}$ & $\mathbf{76.4}$ & $\mathbf{4.2}$\\
         \hline
    \end{tabular}
    }
    \label{tab:moviechat}
    \end{minipage}
    \hfill
    \begin{minipage}{.45\textwidth}
     \caption{\textbf{Performance comparison of short video question answering} on MSRVTT-QA (denoted by MSR-QA), MSVD-QA and ActivityNet-QA (denoted by ANet-QA). The top-1 accuracy in exact matching evaluation setting is reported.}

    \resizebox{\linewidth}{!}{
    \begin{tabular}{l|ccc}
        \hline
         Model & MSR-QA & MSVD-QA & ANet-QA \\
         \hline
         JustAsk \cite{yang2021just} & $41.5$ & $46.3$ & $38.9$\\  
         FrozenBiLM \cite{yang2022zero} &$47.0$ &$54.4$ &$43.2$\\
         SINGULARITY \cite{lei2022revealing}& $43.5$&-& $43.1$\\
         GIT \cite{wang2022git}&$43.2$& $56.8$& -\\
         VIOLETv2 \cite{fu2023empirical} & $44.5$ & $54.7$ & -\\
         mPLUG-2 \cite{xu2023mplug} & $48.0$& $58.1$& -\\
         UMT-L \cite{li2023unmasked} & $47.1$ & $55.2$ & $47.9$\\
         Mirasol3B \cite{piergiovanni2024mirasol3b} & {$50.4$} & - & $51.1$\\
         MA-LMM \cite{he2024ma} & $48.5$&$60.6$&$49.8$\\
         \hline
         \rowcolor[gray]{0.92}
         \textbf{HierarQ$\mathbf{\ddag}$} & $\underline{53.4}$ & $\underline{64.4}$ & $\underline{56.8}$\\
         \rowcolor[gray]{0.92}
         \textbf{HierarQ} & $\mathbf{54.1}$ & $\mathbf{66.2}$ & $\mathbf{57.1}$\\
         \hline
    \end{tabular}
    }
    \label{tab:vqa}
    \end{minipage}
    \vspace{-10pt}
\end{table*}

\noindent \textbf{Medium to Long Video Understanding.} We experiment with three benchmarks: LVU \cite{wu2021towards}, Breakfast \cite{kuehne2014language}, and COIN \cite{tang2019coin}. LVU contains \mytexttilde$30$K videos averaging $1$-$3$ minutes. 
We focus on content understanding across relationship, speaking style, and scene classification tasks. Breakfast includes \mytexttilde$1.7$K videos of breakfast preparation, averaging \mytexttilde$2.7$ minutes, and COIN contains \mytexttilde$11.8$K instructional videos for $180$ tasks with an average duration of \mytexttilde$2.36$ minutes. We report top-1 accuracy for all datasets.

\noindent \textbf{Video Question Answering.} We assess both long and short video QA. For long video QA, we use MovieChat-1k \cite{song2024moviechat}, with $1$K videos averaging \mytexttilde$9.4$ minutes. For short video QA, we evaluate on MSRVTT-QA \cite{xu2017video}, MSVD-QA \cite{xu2017video}, with average durations of $10$-$15$ seconds; and ActivityNet-QA \cite{yu2019activitynet}, with average duration of \mytexttilde$2$ minutes. For MovieChat-1k, we report GPT-3.5-assisted accuracy and scores \cite{maaz-etal-2024-video} to maintain consistency with prior work, while exact matching is used for other QA datasets.

\noindent
\textbf{Video Captioning.} We experiment with three popular video captioning benchmarks: MSRVTT \cite{xu2016msr}, MSVD \cite{chen:acl11}, and YouCook2 \cite{zhou2018towards}, and report CIDEr scores \cite{vedantam2015cider}.

\subsection{Implementation Details}
We use pre-trained ViT G/14 from EVA-CLIP \cite{sun2023eva} as the frozen visual encoder and Vicuna 7B as the frozen LLM. We initialize HierarQ with pre-trained weights from InstructBLIP \cite{dai2023instructblip}. As shown in Figure \ref{fig:main_fig}, we tune the parameters of the feature modulator, HierarQ and FC layer, whereas visual encoder parameters are frozen and LLM parameters are fine-tuned with LoRA \cite{hu2022lora} with rank $32$.
All experiments are conducted on 4 A100 GPUs. 
Additional implementation details have been provided in the supplementary. 

\subsection{Results}
\textbf{Medium to Long Video Understanding.} We present the content understanding results on LVU in Table \ref{tab:lvu}. Notably, our model surpasses the previous best method, MA-LMM \cite{he2024ma}, by $6.8\%$. Additionally, we evaluate our model on the Breakfast and COIN datasets, achieving state-of-the-art performance with improvements of $3.1\%$ and $2.8\%$, respectively, as shown in Table \ref{tab:breakfast}.

\noindent
\textbf{Video Question Answering.} We present the results of our model on long-term video question answering using the MovieChat-1k dataset in Table \ref{tab:moviechat}. Our approach surpasses the current state-of-the-art by $3.5\%$ in Global accuracy and $2.9\%$ in Breakpoint accuracy, while consistently achieving high scores (rated between $1$ to $5$) across both metrics. Additionally, our model demonstrates notable improvements in short-term video question answering tasks, outperforming the state-of-the-art by $3.7\%$, $5.6\%$, and $6\%$ on the MSRVTT-QA, MSVD-QA, and ActivityNet-QA benchmarks, respectively, as shown in Table \ref{tab:vqa}.

\noindent
\textbf{Video Captioning.} To further evaluate free-form text generation capabilities of our model, we conduct experiments on MSRVTT, MSVD and YouCook2 and report their results in Table \ref{tab:vid-caption}. Despite initially being pre-trained with image-text pairs and only fine-tuned with video-text data, our model consistently surpasses the performance of concurrent models some of which contain large-scale video-text pre-training. 

\vspace{-5pt}
\subsection{Ablation Studies}
\label{subsec:ablation}
\vspace{-.5pt}
\textbf{Contribution of each component.} In our ablation studies (Table \ref{tab:ablation}), using the Entity-guided feature modulator alone slightly decreases performance over baseline \cite{he2024ma}, likely due to limited scene context. On the contrary, using the Prompt-guided feature modulator alone improves results by putting more focus on task-relevant features. Introducing HierarQ shows gains with either modulator individually, as it still accesses both entity and scene streams even without modulated features. However, combining all components yields the highest performance, highlighting the value of a hierarchical approach where detailed short-term entity information supports broader long-term scene context.

\begin{table*}[t!]
    \centering
    \small
    \begin{minipage}{.45\textwidth}
    \caption{\textbf{Performance comparison of video captioning} on MSRVTT, MSVD and YouCook2 datasets. The CIDEr scores are reported.}
    \resizebox{\linewidth}{!}{
    \renewcommand{\arraystretch}{1.2}
        \begin{tabular}{l|c|c|c}
        \hline
         Model&MSRVTT&MSVD&YouCook2  \\
         \hline
         SwinBERT \cite{lin2022swinbert} & $55.9$& $149.4$&$109.0$\\
         GIT \cite{wang2022git} & $73.9$&{$180.2$}&$129.8$\\
         mPLUG-2 \cite{xu2023mplug} & $\underline{80.3}$&$165.8$ &-\\
         HowToCaption \cite{shvetsova2023howtocaption}& $65.3$ & $154.2$ &  $116.4$\\
         MA-LMM \cite{he2024ma}&$74.6$ &$179.1$&\underline{$131.2$}\\
         \hline
         \rowcolor[gray]{.92}
         \textbf{HierarQ$\mathbf{\ddag}$} & {$79.8$}&$\underline{182.6}$&$\underline{134.4}$\\
         \rowcolor[gray]{.92}
         \textbf{HierarQ} & $\mathbf{80.5}$&$\mathbf{183.1}$&$\mathbf{136.1}$\\
         \hline
        \end{tabular}
        }
        \label{tab:vid-caption}
\end{minipage}
\hfill
\begin{minipage}{.53\textwidth}
     \caption{\textbf{Contribution of each component.} Here, BL denotes the baseline. Ent., Prt. denote entity and prompt- guided feature modulators (Feat. Mod.). HQ, FT respectively denote HierarQ and LLM Finetuning. Brf denotes Breakfast dataset.}
     \vspace{-1mm}
    \renewcommand{\arraystretch}{.95}
    \resizebox{\linewidth}{!}{
    \begin{tabular}{c|cc|cc|c|c|cc}
        \hline
        \multirow{2}{*}{BL}&\multicolumn{2}{c|}{Feat. Mod.} & \multicolumn{2}{c|}{Memory} & \multirow{2}{*}{HQ} & \multirow{2}{*}{FT} & \multirow{2}{*}{LVU} & \multirow{2}{*}{Brf}\\
        &Ent.&Prt.& Short & Long & & & \\
        \hline
        \checkmark&&&&\checkmark&&&$60.7$&$93.0$\\
        \hline
        \checkmark&\checkmark&&\checkmark&&&&$58.7$&$88.5$\\
        \checkmark&&\checkmark&&\checkmark&&&$62.0$&$94.1$\\
        \hline
        \checkmark&&\checkmark&\checkmark&\checkmark&\checkmark&&$65.7$&$95.2$\\
        \checkmark&\checkmark&&\checkmark&\checkmark&\checkmark&&$65.0$&$94.8$\\

        \rowcolor[gray]{0.92}
        \hline
        \checkmark&\checkmark&\checkmark& \checkmark&\checkmark&\checkmark&&$\underline{66.8}$&$\underline{96.1}$\\
        \rowcolor[gray]{0.92}
        \checkmark&\checkmark&\checkmark& \checkmark&\checkmark&\checkmark&\checkmark&$\mathbf{67.9}$&$\mathbf{97.4}$\\
        \hline
    \end{tabular}}
    \label{tab:ablation}
    \end{minipage}
    \vspace{-3.5mm}
\end{table*}

\begin{table}[t!]
    \centering
    \small
    \caption{\textbf{Contribution of visual and query memory} at short and long-term memory banks. FT denotes LLM finetuning.}
    \renewcommand{\arraystretch}{.9}
    \begin{tabular}{cc|cc|c|cc}
        \hline
        \multicolumn{2}{c|}{Short-term}&\multicolumn{2}{c|}{Long-term}&\multirow{2}{*}{FT}&\multirow{2}{*}{LVU}&\multirow{2}{*}{Breakfast}\\
        Visual & Query &Visual & Query&\\
        \hline
        &&&&&$55.8$&$75.3$\\
        \hline
        \checkmark&&&&&$61.8$&$88.2$\\
        &\checkmark&&&&$59.3$&$81.3$\\
        \checkmark&\checkmark&&&&$62.7$&$90.4$\\
        \hline
        &&\checkmark&&&$62.5$&$90.9$\\
        &&&\checkmark&&$61.6$&$88.3$\\
        &&\checkmark&\checkmark&&$64.9$&$91.8$\\
        \hline
        \checkmark&\checkmark&\checkmark&&&$65.1$&$93.3$\\
        \checkmark&\checkmark&&\checkmark&&$63.6$&$92.1$\\
        \hline
        \checkmark&&\checkmark&\checkmark&&$65.4$&$94.2$\\
        &\checkmark&\checkmark&\checkmark&&$62.0$&$93.8$\\
        \rowcolor[gray]{0.92}
        \hline
        \checkmark&\checkmark& \checkmark&\checkmark&&$\underline{66.8}$&$\underline{96.1}$\\
        \rowcolor[gray]{0.92}
        \checkmark&\checkmark& \checkmark&\checkmark&\checkmark&$\mathbf{67.9}$&$\mathbf{97.4}$\\
        \hline
    \end{tabular}
    \label{tab:memory-bank-abla}
    \vspace{-15pt}
\end{table}

\vspace{-2pt}
\noindent
\textbf{Contribution of each memory bank.} Our ablation studies (Table \ref{tab:memory-bank-abla}) demonstrate the importance of both short- and long-term memory banks in providing rich temporal context. Without memory banks, performance declines significantly, highlighting their critical role in temporal modeling. The visual memory bank, which stores raw historical features, offers a clearer temporal context and thus better performance, compared to the query memory bank, which encodes information more implicitly. Long-term memory alone outperforms short-term memory, as it captures essential broader dynamics. However, the combination of both short- and long-term memory banks—each containing visual and query memories—achieves the best performance. This is likely due to the long-term memory’s ability to provide broader context, while short-term memory complements it by adding frame-specific information. This hierarchical structure, where short-term memory enhances long-term scene understanding, proves highly effective.

\begin{figure*}
    \centering
    \begin{minipage}{.4\textwidth}
        \includegraphics[width=\linewidth, height=4cm]{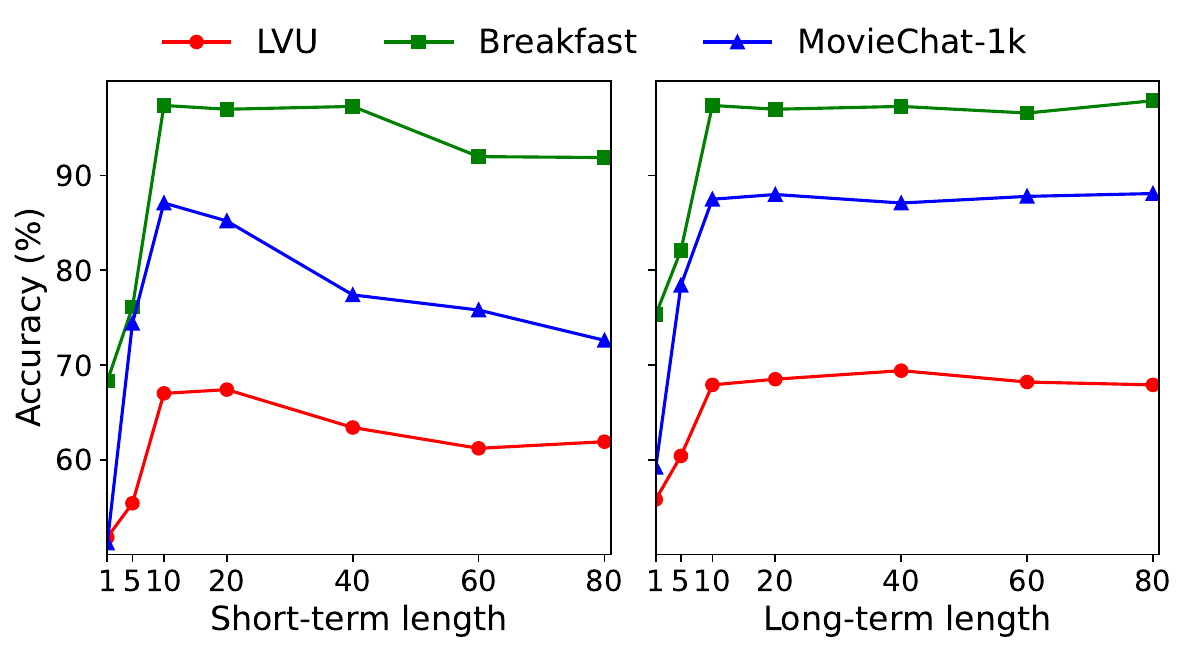}
        \caption{\textbf{Impact of memory bank length.} When one memory length is varied, the other one remains at a fixed size of $10$. 
        }
        \label{fig:mb-length}
    \end{minipage}
    \hspace{3pt}
    \begin{minipage}{.58\textwidth}
        \includegraphics[width=.98\linewidth]{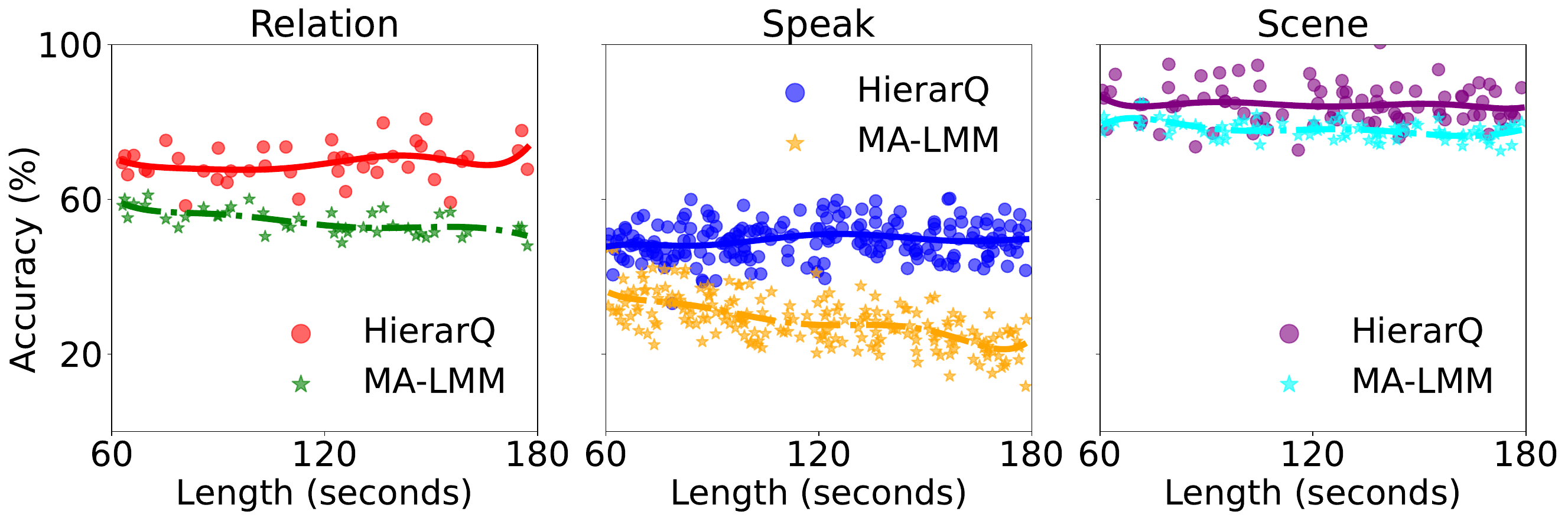}
        \caption{\textbf{Impact of video length.} Here, across both relation and speak category of LVU dataset, MA-LMM performance decreases as video length increase, however, HierarQ achieves fairly stable performance showing the effectiveness of our method across increasing video length.}
        \label{fig:len_acc}

    \end{minipage}
    \vspace{-5pt}
\end{figure*}

\noindent
\textbf{Memory bank length ablation.} In Figure \ref{fig:mb-length}, we analyze how varying memory bank lengths impact model accuracy at the short and long term. Initially, increasing memory length enhances accuracy, as it provides richer historical context. However, performance for the short-term memory decreases beyond a length of $10$, likely due to excess entity-focused information overshadowing scene-level context, reducing its complementary role. For long-term memory, performance plateaus after a length of $10$, indicating that retaining all historical data isn’t necessary due to temporal redundancy. These findings suggest that a balanced memory length between entity and scene level optimally supports long-term video comprehension.

\begin{figure*}
    \centering
    \begin{minipage}{.41\textwidth}
        \includegraphics[width=\linewidth]{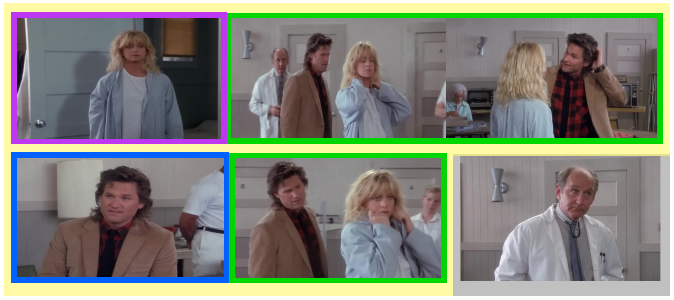}
        \caption{\textbf{Visualization of feature-modulated frames.} For the prompt \texttt{"What is the speaking style of the man and woman?"}, blue, purple, and green borders show frames emphasized by the entity modulator, while yellow shading highlights scene-relevant frames. Gray regions denote irrelevant frames.
        }
        \label{fig:qual}
    \end{minipage}
    \hspace{3pt}
    \begin{minipage}{.57\textwidth}
        \vspace{-3pt}
        \includegraphics[width=.33\linewidth]{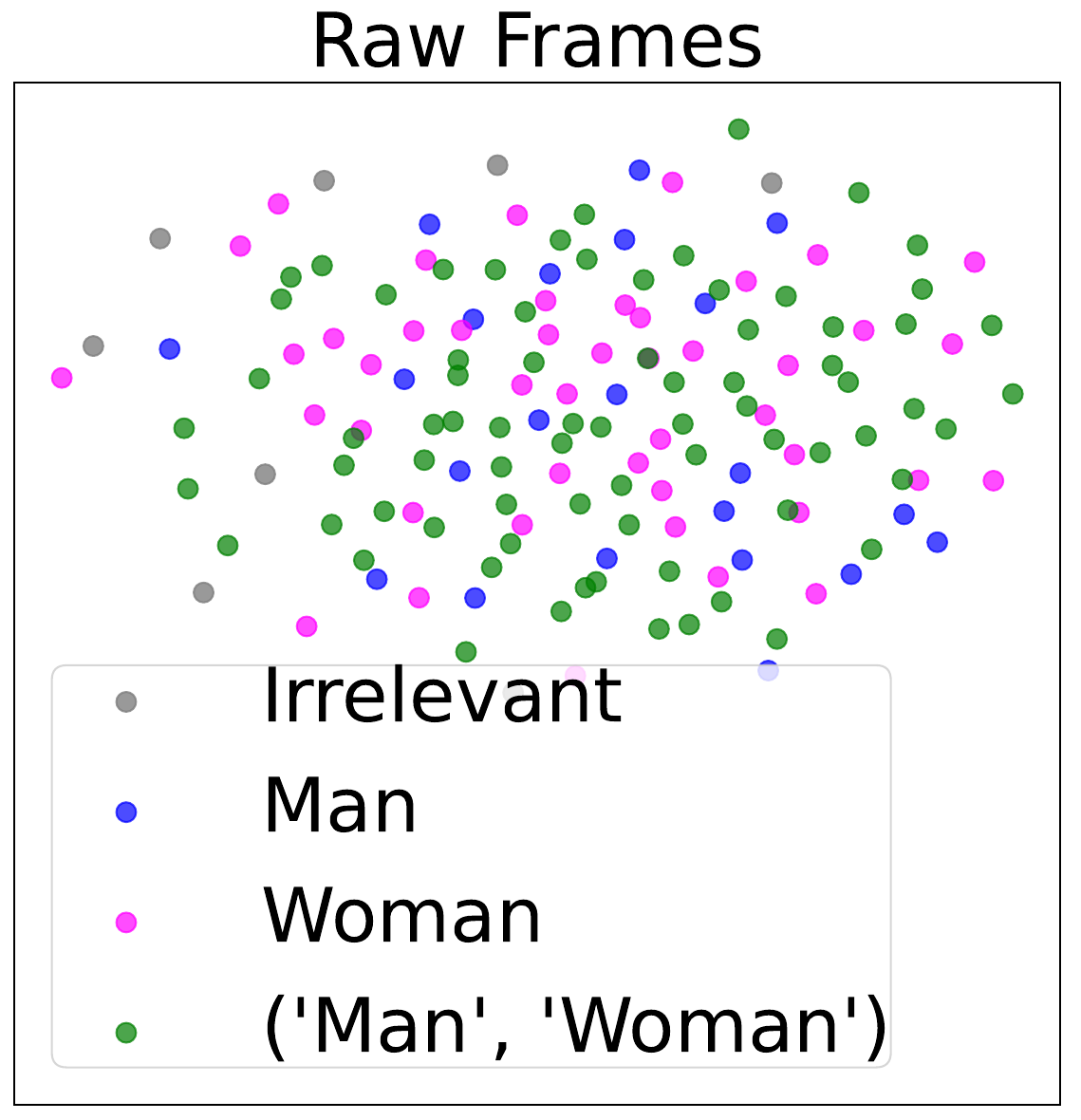}
        \includegraphics[width=.32\linewidth, height=3.4cm]{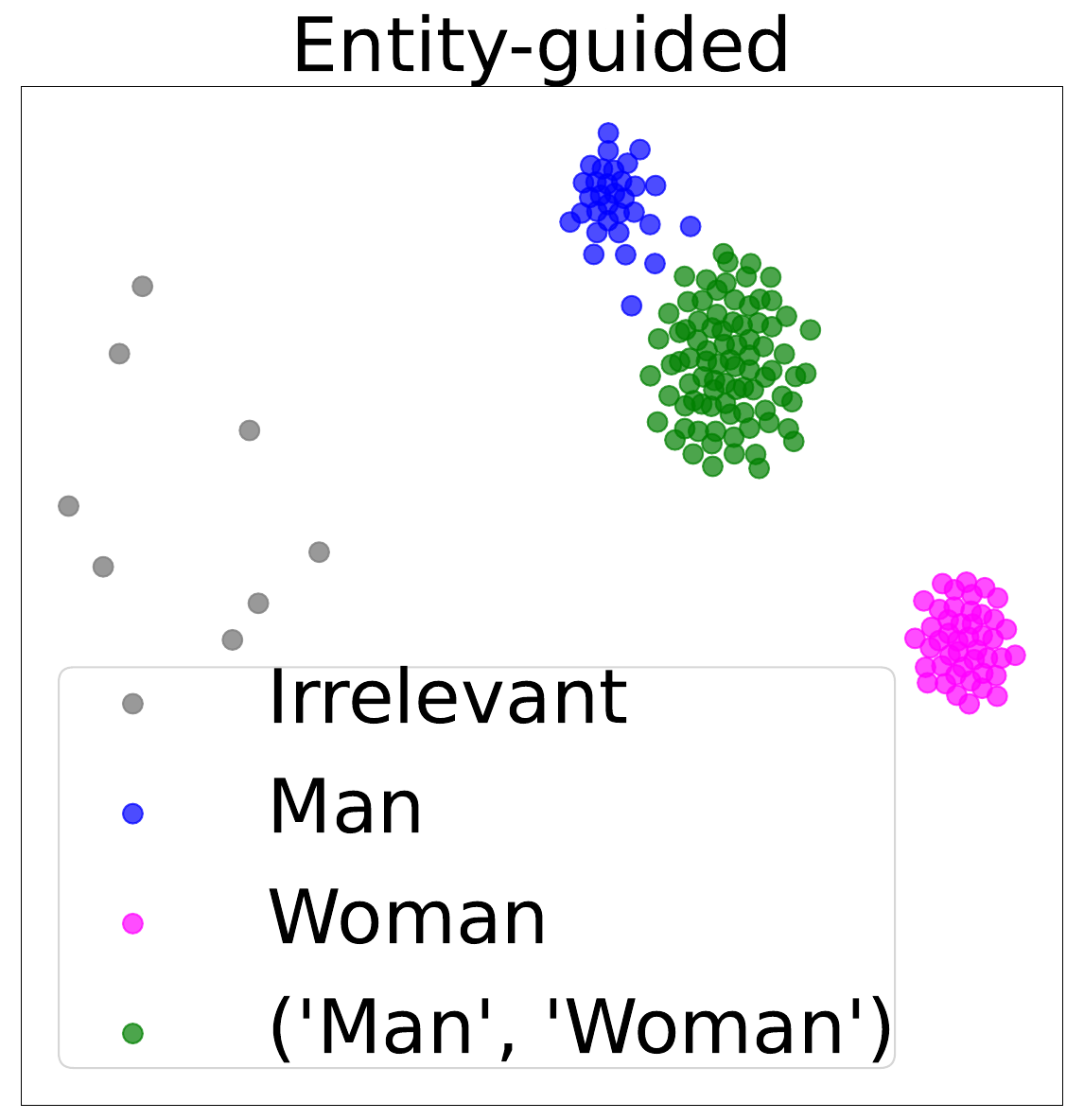}
        \includegraphics[width=.33\linewidth]{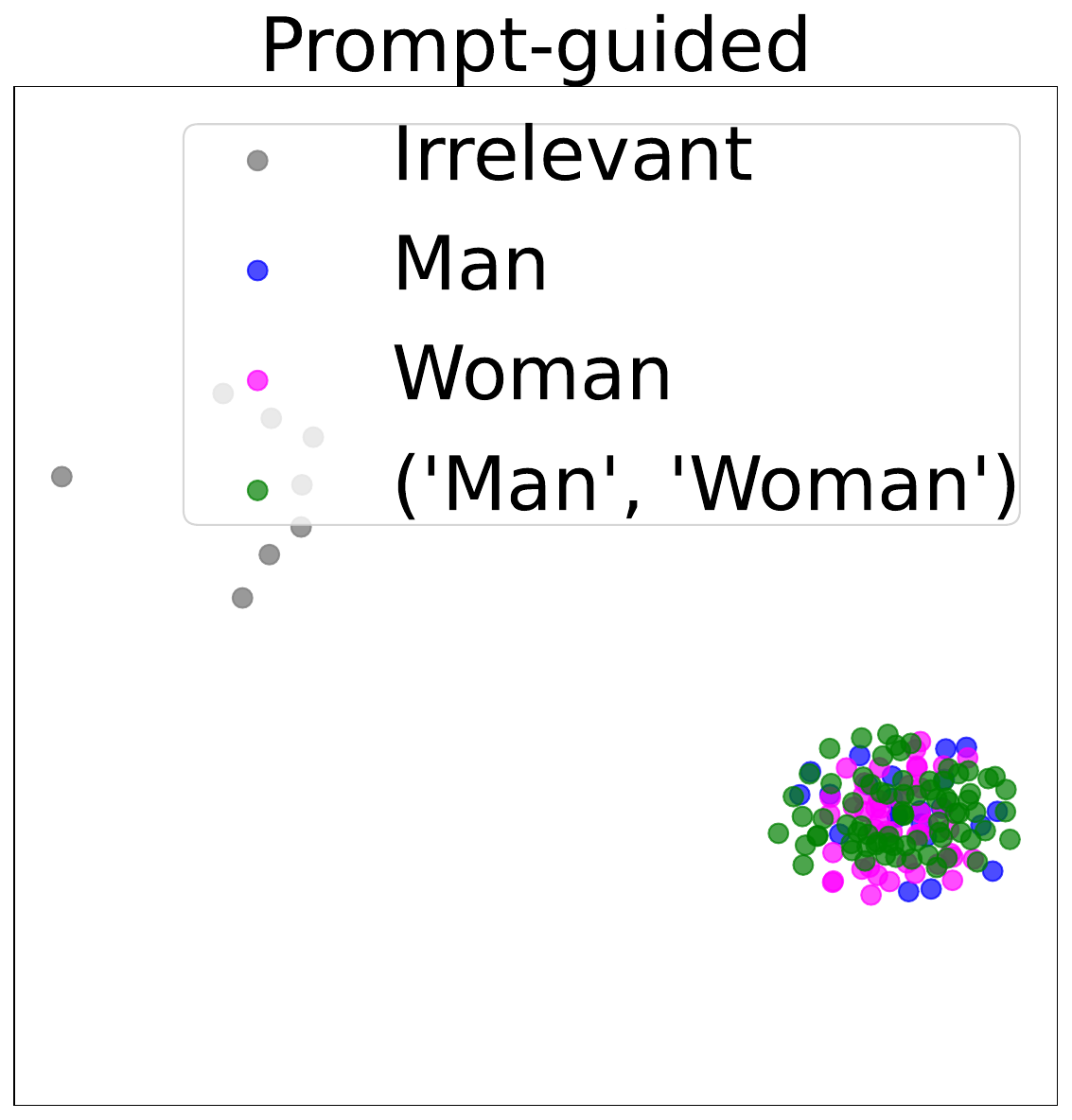}
        \caption{
        \textbf{Effect of feature modulators on feature space.} The t-SNE plot shows distinct clustering of frames after processing through entity-guided and prompt-guided feature modulators, highlighting their effectiveness in focusing on relevant details for enhanced video understanding. Cluster pseudo-labels are assigned based on the inspection of few representative frames from each visually distinct clusters.
        }

        \label{fig:tsne}
    \end{minipage}
    \vspace{-10pt}
\end{figure*}

\noindent
\textbf{Memory update method.} Our ablation studies (Table \ref{tab:memory-compression}) reveal key insights into optimal memory design choices for our framework. Using FIFO for both short and long-term memory results in moderate performance due to the limited historical context. Switching both to retain historical information with memory bank consolidation (MBC), significantly boosts performance, as both memory banks now retain a longer temporal context. However, since the short-term memory contains frame-specific entity details, we hypothesize that it benefits less from extended context, whereas long-term memory requires historical data for understanding entity interactions over time. The best performance is achieved by combining FIFO for short and MBC for long-term memory, confirming that short-term memory works best with a focused, short-term context, while long-term memory excels with a broader temporal scope.

\begin{table}[t!]
    \centering
    \caption{\textbf{Impact of different memory update methods.}}
    \begin{tabular}{cc|cc}
        \hline
         Short-term &Long-term & LVU & Breakfast \\
         \hline
         FIFO & FIFO & $65.2$& $93.6$ \\
         MBC & MBC & $67.4$ & $97.3$ \\
         \hline
         \rowcolor[gray]{0.92}
         FIFO & MBC & $\mathbf{67.9}$& $\mathbf{97.4}$\\
         \hline
    \end{tabular}
    \vspace{-14pt}
    \label{tab:memory-compression}
\end{table}

%% file: sec/5_analysis.tex
\vspace{-6pt}
\section{Analysis and Discussion}
\label{sec:analysis}
\textbf{How effective is HierarQ when video length increases?} To demonstrate effectiveness across varying video lengths, we compare our model with MA-LMM \cite{he2024ma} in Figure \ref{fig:len_acc}. Results indicate that as video length increases, MA-LMM’s performance declines in both relation and speaking style classification categories, highlighting the need for effective long-context modeling in these tasks. In contrast, our model maintains stable performance across categories regardless of video length, confirming its superior capability for handling extended contexts.

\noindent \textbf{Do we need hierarchical relationship modeling?} To demonstrate that HierarQ effectively models long-term information with broad context understanding of the scene-level Q-Former while receiving complementary short-term cues from the entity-level Q-Former, we conducted an ablation by isolating the two Q-Formers. In this setup, each Q-Former learns independently without interacting with the other. We then concatenated the learned queries from both streams before sending them to the FC layer. This variant resulted in a performance drop of $3.6$\% on the LVU dataset, confirming that the hierarchical structure in HierarQ significantly enhances the scene-level Q-Former’s learning with help of the entity-level Q-Former and supports comprehensive video understanding.

\noindent \textbf{Is it just more parameters?} To ensure HierarQ’s gains aren’t solely due to additional parameters, we perform an ablation adding additional layers to the baseline Q-Former \cite{he2024ma} to match HierarQ’s parameter count. This variant, lacking dual branches for short- and long-term memory, 
underperforms by $4.7$\% compared to HierarQ. This confirms that HierarQ’s architecture, rather than just added parameters, drives its capacity for capturing complex relationships.

\vspace{-2pt}
\subsection{Qualitative Analysis}
In Figure \ref{fig:qual}, we present frames from a video of LVU dataset, showcasing the impact of our two-stream feature modulators. After modulation, entity and prompt-relevant frames are distinctly clustered (Figure \ref{fig:tsne}), with entity- and prompt-focused frames distributed throughout the video timeline. The prompt-guided feature modulator effectively focuses on all task-relevant frames while putting less emphasis on irrelevant ones, resulting in a broader scene-level understanding that is enriched by entity-level details. This clustering highlights our model’s ability to emphasize both entity-specific and scene-wide information, achieving comprehensive video understanding. More qualitative analysis have been provided in the supplementary material.

%% file: sec/6_conclusion.tex
\vspace{-7pt}
\section{Conclusion}
\label{sec:conclusion}

In this paper, we introduce \textbf{HierarQ}, a novel task-aware hierarchical querying transformer based framework designed for effective video understanding over extended time periods. Our framework processes video frames sequentially, addressing limitations related to LLM's context length and input frame capacity. To incorporate task-awareness in the processing method, our approach includes a two-stream feature modulator with an entity-guided branch for frame-level object understanding within a short context, and a prompt-guided branch for broader scene understanding over a longer context, capturing interactions among entities. These streams are supported by dedicated memory banks that balance immediate details and broader context, supported by our \textbf{HierarQ} (\textbf{Hi}erarchical \textbf{Q}uerying transformer) for enhanced scene comprehension by integrating entity-level details. Extensive evaluations on 10 popular benchmarks for video understanding, question answering, and captioning highlight HierarQ’s state-of-the-art performance across most of these datasets and competitive performance on others,
demonstrating its robustness and adaptability.

\section{Acknowledgement}
This research has benefitted from the Microsoft Accelerating Foundation Models Research (AFMR) grant program.

%% file: sec/X_suppl.tex
\clearpage
\setcounter{page}{1}
\renewcommand{\thesection}{\Alph{section}}
\setcounter{section}{0}

\maketitlesupplementary

In this supplementary material, we provide additional quantitative results in Section \ref{sec:res_sup}. Then we present additional ablation studies in Section \ref{sec:ana_sup}, followed by qualitative analysis in Section \ref{sec:res_qual}. Next, we provide more implementation details in Section \ref{sec:imp_sup}. Finally, we address some limitations and outline directions for future research in Section \ref{sec:future}.

\section{Additional Results}
\label{sec:res_sup}
In Table \ref{tab:lvu_meta} we present the results for metadata classification task on the LVU dataset. The metadata classification task consists of - director, genre, writer, and year categories. Similar to the content understanding task presented in Table \ref{tab:lvu}, HierarQ achieves state-of-the-art performance with an average performance gain of $5.5$\% over the best model \cite{he2024ma} proving its the effectiveness across multiple categories of classification task. Additionally, in Table \ref{tab:vid-caption-meteor}, we present performance comparison of our model against other concurrent approaches on the task of video captioning and report the METEOR scores.

\section{Additional Analysis}
\label{sec:ana_sup}

\begin{figure*}[t!]
    \centering
    \includegraphics[width=.8\linewidth]{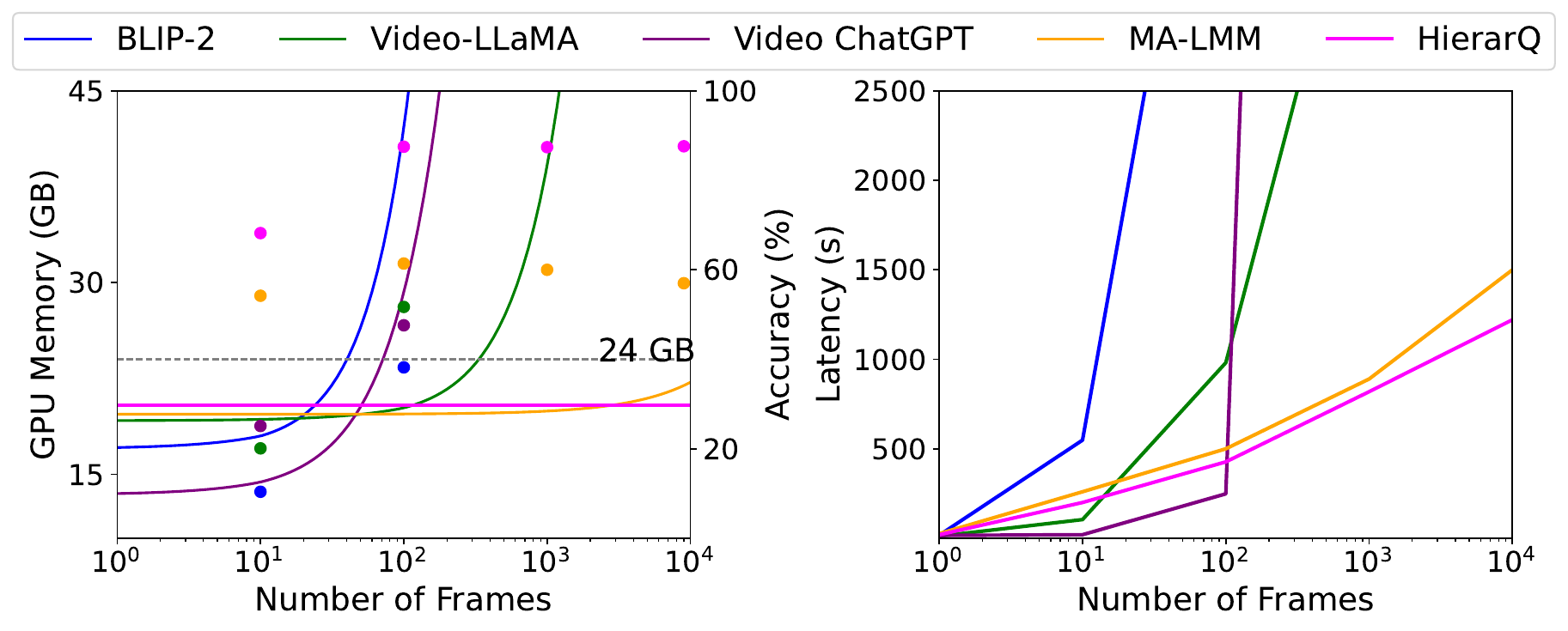}
    \caption{\footnotesize GPU memory (\textit{line}), accuracy (\textit{dots}) and latency vs. frames.}
    \label{fig:merged}
    \vspace{-10pt}
\end{figure*}

\textbf{Computational Cost Analysis.} In Figure \ref{fig:merged} we present detailed computation cost analysis of our method HierarQ and compare it against baseline Q-Former (BLIP-2 \cite{li2023blip}) and few other concurrent approaches. Token: HierarQ uses a $32$-token count, matching the token-count of MA-LMM and Video LLaMA while being smaller in tokens than baseline Q-Former (BLIP-2) and Video ChatGPT. Memory bank size: HierarQ uses fixed-size memory banks with $10$ frames and $32$ tokens per frame for both short- and long-term memory bank. In contrast, MA-LMM uses a variable-sized memory ranging from $10$ to $40$ frames with $32$ tokens per frame, while MovieChat has $18$ frames and $32$ tokens per frame in short-term memory and $256$ frames in long-term memory. Computation cost: HierarQ maintains constant memory consumption, processing over $10$K frames on a $24$GB A$100$ GPU (Figure \ref{fig:merged} left). Unlike models with exponential computation growth, its auto-regressive design ensures scalability, fitting within a $24$GB GPU for $100$-frame inputs. Its training-free memory banks incur no additional computation costs, while the number of trainable parameters is only $390$M. Latency: While latency increases linearly with frames (Figure \ref{fig:merged} right), HierarQ remains capable of processing arbitrarily long videos, unlike other models that fail beyond a certain frame threshold.

\noindent
\textbf{Temporal modeling method ablation.} Table \ref{tab:temp-model} compares different temporal modeling methods. Some strategies to reduce token load due to auto-regressive frame processing is to do concatenation, average pooling or even token merging (ToMe) \cite{bolya2022token}. Our HierarQ outputs $32$ tokens per frame, with simpler methods like concatenation and averaging of frame features yielding lower performance. Concatenation, in particular, is computationally expensive due to simultaneous processing of all frames. ToMe reduces tokens per frame from $32$ to $2$, but for $100$-frame inputs, it still requires $200$ tokens, imposing significant memory demands and sub-optimal performance. In contrast, our framework utilizes entity and scene-level task-aware streams with dedicated memory banks to store historical temporal information. The short-term memory bank employs a FIFO approach, while the long-term memory bank performs compression of similar token based on high cosine similarity (MBC). This temporal modeling technique keeps the token count fixed at $32$ per frame, while not losing essential information and also optimizing GPU memory usage. Our temporal modeling approach achieves superior accuracy on the LVU and Breakfast datasets as compared to the other approaches. Here, we limit the number of frames to $100$ for a fair comparison to the other methods that risk facing the LLM context length bottleneck with increased frames. Here, it is important to mention that previous works \cite{song2024moviechat, he2024ma} have successfully performed temporal modeling respectively using ToMe and MBC. However, here we empirically show that our combination of using short and long term memory for temporal modeling gives the best performance.

\begin{table}[t]
\centering
\footnotesize
    \caption{\textbf{Performance comparison of medium to long video understanding} on LVU dataset. The top-1 accuracy is reported. $\ddag$ indicates without LLM finetuning. \textbf{Best} and \underline{second-best} performances are highlighted.}
    \label{tab:lvu_meta}
    \begin{tabular}{l|cccc|c}
    \hline
    Model & Director & Genre & Writer & Year & Avg \\
    \hline
    VideoBERT   \cite{sun2019videobert}           & $47.3$      & $51.1$   & $38.5$   &   $36.1$ & $43.3$ \\
    Obj\_T4mer\cite{wu2021towards}            & $47.7$      & $52.7$   & $36.3$  &  $37.8$ &  $43.6$                                           \\
    Orthoformer   \cite{patrick2021keeping}         & $55.1$      & $55.8$   & $47.0$   &    $43.4$  & $50.3$                                        \\

    VIS4mer   \cite{islam2022long}             & $62.6$      & $54.7$   & $48.8$   &                                       $44.8$      & $52.7$\\

    TranS4mer \cite{islam2023efficient} & $63.9$ & $55.9$ & $46.9$ & $45.5$ & $53.1$\\
    S5 \cite{wang2023selective}                    & $67.3$ & $65.4$      & $51.3$   & $48.0$   &  $58.0$                                           \\
    Movies2Scene  \cite{chen2023movies2scenes} & $70.9$& $55.9$&$53.7$&$57.8$ & $59.6$\\
    VideoMamba \cite{li2025videomamba} & $67.3$ & $65.2$ & $53.0$& $48.2$ & $58.4$\\
    MA-LMM  \cite{he2024ma}               & $74.6$      & $61.1$   & $70.4$   &$51.9$  & $64.5$                                            \\
    \hline
    \rowcolor[gray]{0.92}
    \textbf{HierarQ $\mathbf{\ddag}$}                   &     \underline{$76.6$}      & $\underline{66.2}$       & $\underline{71.1}$       & $\underline{59.2}$  &$\underline{68.3}$ \\ 
    \rowcolor[gray]{0.92}
    \textbf{HierarQ}                   &     $\mathbf{78.4}$      & $\mathbf{67.9}$       & $\mathbf{71.9}$       & $\mathbf{61.9}$ & $\mathbf{70.0}$    \\ 
    \hline
    \end{tabular}
\end{table}

\begin{table}[t!]
    \centering
    \small

    \caption{\textbf{Performance comparison of video captioning.} Here we report the METEOR scores.}

    \begin{tabular}{l|c|c|c}
        \hline
         Model&MSRVTT&MSVD&YouCook2  \\
         \hline
         SwinBERT \cite{lin2022swinbert}  & $29.9$& $41.3$&$15.6$\\
         GIT \cite{wang2022git} & $32.9$&\underline{$51.1$}&$17.3$\\
         mPLUG-2 \cite{xu2023mplug}& $34.9$&$48.4$ &-\\
         HowToCaption \cite{shvetsova2023howtocaption}& $32.2$ & $46.4$ &  $15.9$\\
         MA-LMM \cite{he2024ma}&\underline{$33.4$} &$51.0$&\underline{$17.6$}\\
         \hline
         \rowcolor[gray]{.92}
         \textbf{HierarQ} & $\mathbf{35.1}$&$\mathbf{51.2}$&$\mathbf{18.1}$\\
         \hline
        \end{tabular}

        \label{tab:vid-caption-meteor}
\end{table}
\noindent
\textbf{Long-term memory bank compression at different spatial levels.} Table \ref{tab:mem-level} compares the performance of compressing the long-term memory bank at different spatial levels: frame-level and token-level, on the LVU and Breakfast datasets. In frame-level compression, cosine similarity is calculated between adjacent frame features, and features with the highest similarity are averaged. In token-level compression, cosine similarity is computed between tokens at the same spatial location across the entire temporal axis, leveraging the fact that each frame feature comprises multiple spatial tokens. We hypothesize that token-level compression preserves finer spatial details compared to frame-level compression. Proving our hypothesis, the results demonstrate that token-level compression consistently outperforms frame-level compression, supporting its ability to retain more detailed spatial information. 

\begin{table}[t!]
    \centering
    \small
    \caption{\textbf{Ablation of different temporal modeling techniques.}}
    \begin{tabular}{c|ccc|cc}
        \hline
         Method & \#Frame & \#Token & GPU & LVU & Breakfast \\
         \hline
         Concat & 60 & 1920 & 53.1 & 65.1 & 94.2  \\
         Avg Pool & 100 & 32 & 25.9& 60.3 & 84.0\\
         ToMe \cite{bolya2022token} & 100 & 200 & 26.6 & 65.6 & 95.7\\
         \hline
         \rowcolor[gray]{0.92}
         \textbf{Ours} & 100 & 32 & 22.4 &$\mathbf{67.9}$& $\mathbf{97.4}$\\
         \hline
    \end{tabular}
    \label{tab:temp-model}
\end{table}

\begin{table}[t!]
    \centering
    \caption{\textbf{Long-term memory bank compression strategy.}}
    \label{tab:mem-level}
    \begin{tabular}{c|cc}
        \hline
         Spatial Level& LVU & Breakfast \\
         \hline
         Frame-level &63.5&94.8\\
         \rowcolor[gray]{0.92}
         Token-level & $\mathbf{67.9}$ & $\mathbf{97.4}$\\
         \hline
    \end{tabular}
\end{table}

\begin{figure*}
    \centering
    \includegraphics[width=\linewidth]{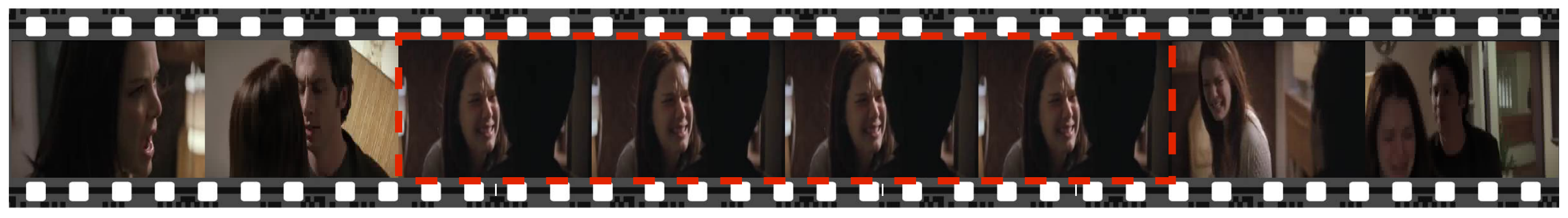}
    \includegraphics[width=\linewidth]{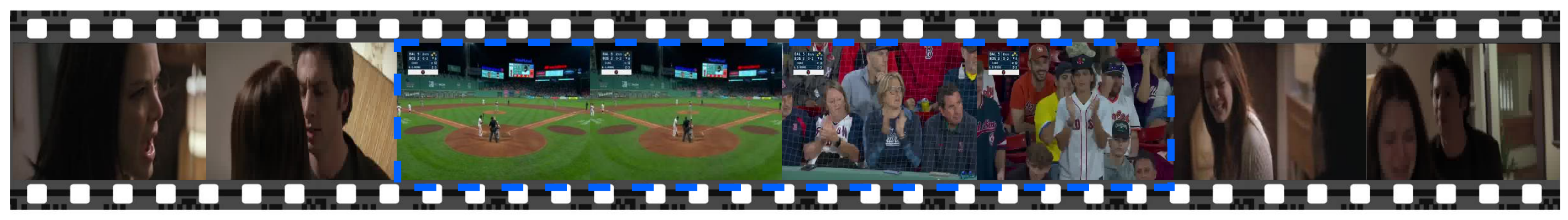}
    \caption{\textbf{Sample frames from original video with irrelevant frames synthetically introduced to it} for evaluating the effectiveness of the task-aware feature modulator. Here, the \textit{top} sample shows the frozen frames setup and the \textit{bottom} sample shows the OOD frames setups. The \textcolor{red}{red} and \textcolor{blue}{blue} boxes respectively denote the frozen and OOD frames.}
    \label{fig:ood}
\end{figure*}
\begin{figure*}
    \centering
    \includegraphics[width=\linewidth]{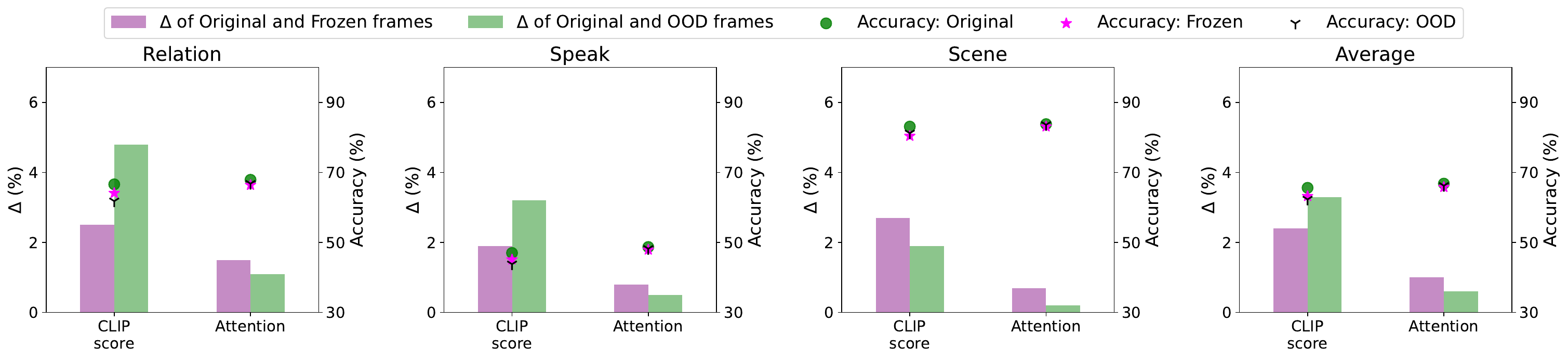}
    \caption{\textbf{Ablation of different feature modulation methods} on LVU dataset. Here, \textit{left} y axis shows accuracy drop ($\Delta$) between original and frozen/OOD frames and \textit{right} y axis shows accuracy of original, frozen and OOD frames. Across all categories the attention  mechanism bridges the performance gap more effectively than CLIP scoring while consistently getting high accuracy showing the effectiveness of the selected feature modulation method.}
    \label{fig:ft-relev-abla}
\end{figure*}

\noindent
\textbf{Feature modulation method ablation.} To assess the task-aware feature modulation method's effectiveness, we conducted experiments using a randomly sampled subset of the LVU dataset's content-understanding videos, constituting $50$\% of the test set. We perform evaluation in two setups: frozen frames and out-of-distribution (OOD) frames. In the frozen frames setup, a single frame from each clip is repeated $10$ times, testing the modulator's ability to identify task-relevant frames despite redundancy. In the OOD frames setup, $10$ frames in each clip are replaced with frames from a randomly chosen sports video from YouTube \footnote{\href{https://www.youtube.com/watch?v=VUZPY2biqJU&pp=ygUJYmFzZWJhbGwg}{Youtube link of OOD sample.}}, simulating irrelevant content to evaluate the modulator's filtering capabilities. Samples for both setups are shown in Figure \ref{fig:ood}.

We compare two modulation methods: CLIP scoring \cite{radford2021learningtransferablevisualmodels} and multi-headed attention. In the CLIP-based approach, frame relevance is scored and weighted against the text prompt, with higher scores indicating stronger relevance. As shown in Figure \ref{fig:ft-relev-abla}, the attention-based method outperforms CLIP scoring by better bridging the performance gap in both setups, demonstrating superior efficiency in filtering task-relevant frames and maintaining high performance across LVU dataset's all content understanding categories.

\begin{figure}
    \centering
    \includegraphics[width=\linewidth]{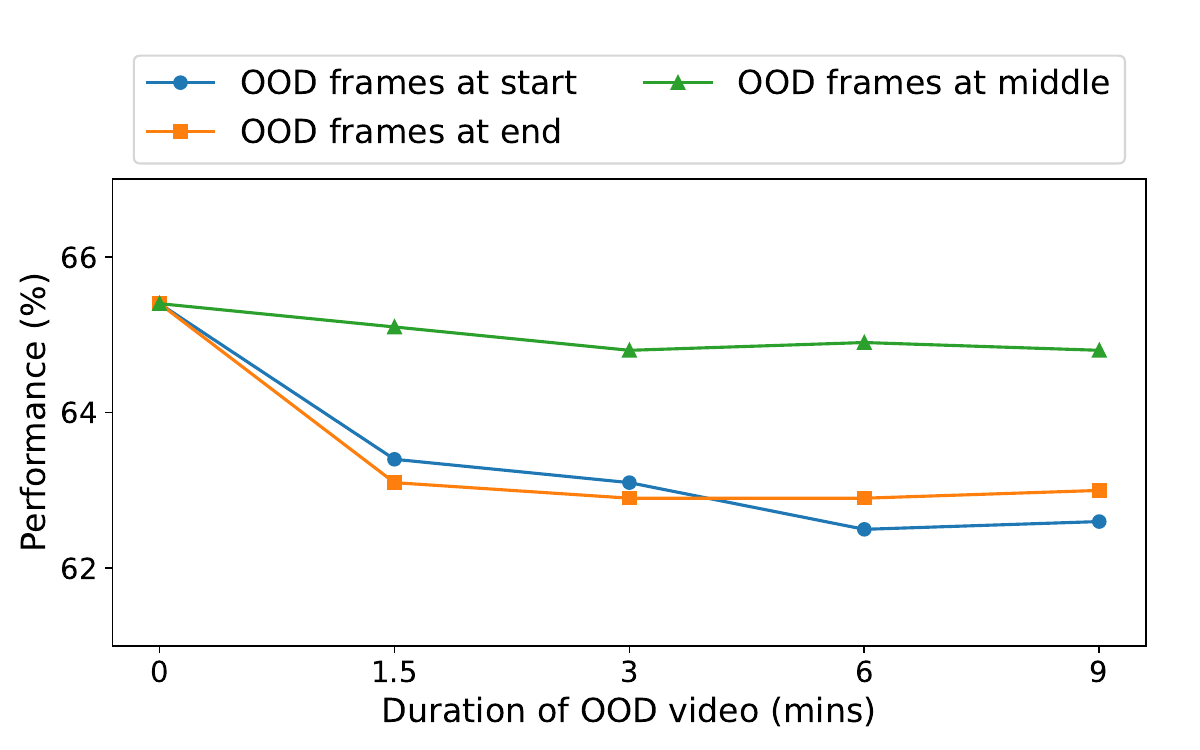}
    \caption{\textbf{Effect of increasing the number of irrelevant frames.} Here, y-axis denotes performance with irrelevant (OOD) frames inserted in different variations and x-axis denotes duration of OOD video where $0$ denotes no OOD frames being inserted, e.g. original video.}
    \label{fig:irrelevant-len}
\end{figure}

\noindent
\textbf{Effect of increasing the number of irrelevant frames.} To further evaluate the impact of irrelevant frames on performance, we conducted experiments using the same randomly sampled subset of the LVU dataset's content-understanding videos as before.
We maintained the original video length while introducing out-of-distribution (OOD) frames to extend the total duration of the video. We added OOD frames under four duration setups: $1.5$ minutes, $3$ minutes, $6$ minutes and $9$ minutes. These OOD frames were inserted in three variations: at the beginning, at the end, and in the middle of the video. 

As shown in Figure \ref{fig:irrelevant-len}, the performance drop saturates even with an increasing number of OOD frames. This indicates that our feature modulator effectively filters out irrelevant information, thus helping HierarQ to maintain high performance despite the increased irrelevant video length. Furthermore, the position of OOD frames (beginning, middle, or end) shows neither significant nor conclusive impact on performance, demonstrating the robustness of our framework in modeling long-term temporal relationships irrespective of the position of irrelevant frames.  

\begin{figure}
    \centering
    \includegraphics[width=\linewidth]{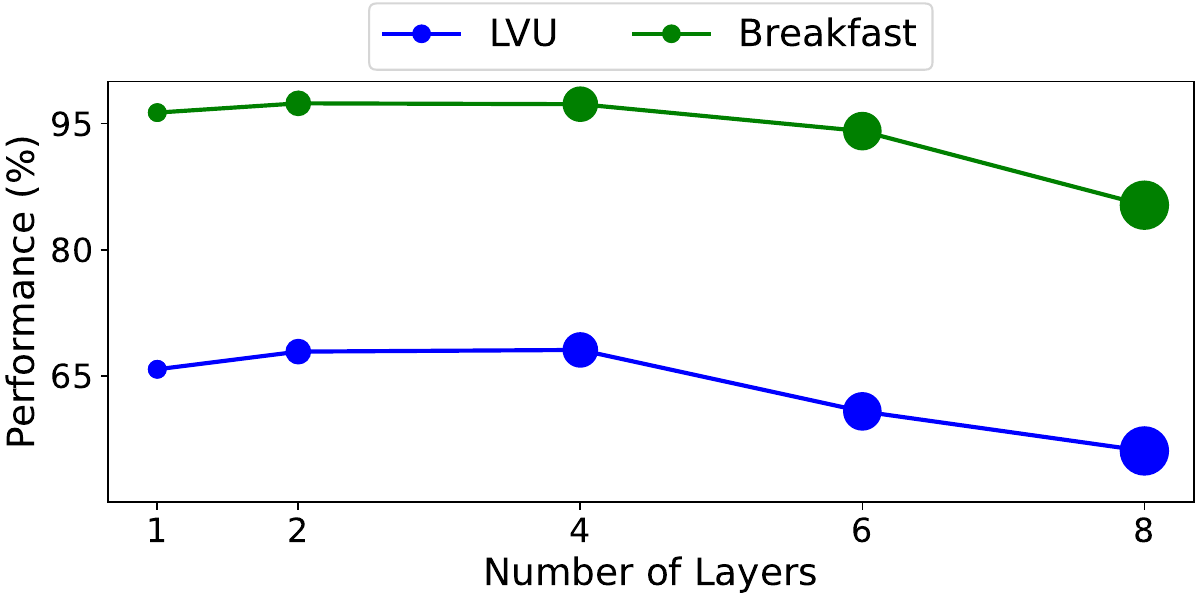}
    \caption{\textbf{Effect of number of layers in the feature modulators.} Here, the marker size denotes the number of parameters.}
    \label{fig:ca-layer-ab}
\end{figure}
\noindent
\textbf{Effect of number of layers in the feature modulator.} To assess the impact of varying the number of cross-attention layers in the task-aware two-stream feature modulators, we present the ablation results in Figure \ref{fig:ca-layer-ab}. The results indicate that increasing the number of layers initially improves performance, as more layers can model nuanced relationships effectively. Performance peaks at $2$ layers for both LVU and Breakfast datasets, suggesting this is the optimal point for capturing task-relevant interactions without overfitting. Beyond $2$ layers, performance plateaus and declines after $4$ layers due to overfitting and diminishing returns, especially as parameter count increases.  The choice between $2$ and $4$ layers presents a trade-off: while $4$ layers offer a marginal improvement on LVU, the linear increase in parameter count makes this configuration less efficient. Therefore, we adopt the $2$-layer architecture as it strikes the best balance between performance and computational efficiency.

\noindent
\textbf{Choice of LLM.} Our model supports various LLM architectures. To identify the best performer, we evaluated three popular LLMs: FlanT5-XL \cite{chung2024scaling}, LLaMA-2 \cite{touvron2023llama2}, and Vicuna-7B \cite{zheng2023judging}. As shown in Table \ref{tab:llm-ablation}, Vicuna-7B achieves slightly better performance than the others.
\begin{table}[t!]
    \centering
    \caption{\textbf{Performance comparison of different LLMs.} Here we report the top-1 accuracy for LVU and Breakfast and global accuracy for MovieChat-1k.}
    \begin{tabular}{cc|ccc}
        \hline
         LLM & Size & LVU&Breakfast&MovieChat-1k \\
         \hline
         FlanT5-XL & 3B & $66.0$ & $95.2$ & $86.7$ \\
         LLaMA-2 & 7B&$67.3$&$97.1$&$87.2$\\
         \rowcolor[gray]{0.92}
         \hline
         Vicuna &7B &$\mathbf{67.9}$&$\mathbf{97.4}$&$\mathbf{87.5}$\\
         \hline
    \end{tabular}
    \label{tab:llm-ablation}
\end{table}

\begin{figure*}[t!]
    \centering
    \includegraphics[width=.85\linewidth]{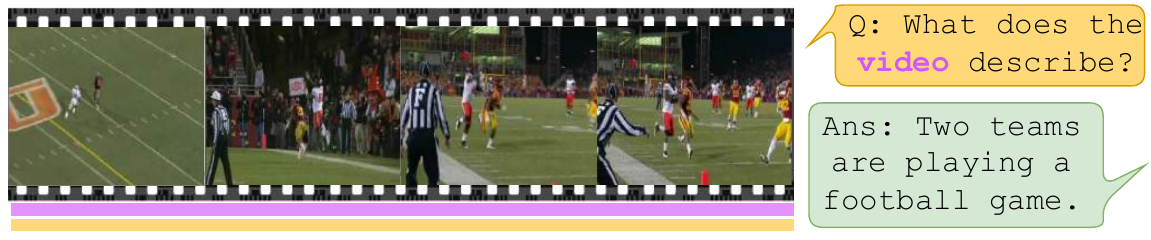}
    \caption{\textbf{Qualitative analysis of video captioning} on MSRVTT. Here, generic nouns (e.g., \textit{``video"}) serve as entity and thus the entity guided feature modulator highlights the entire video. 
    }
    \label{fig:vidcap}
\end{figure*}

\noindent
\textbf{Robustness of entity guided feature modulator.} 
Entities are extracted using POS tagging (\textit{NN, NNS, NNP, NNPS}) with Python's NLTK library, covering all common and proper nouns. HierarQ uses long-term relationship modeling to focus on relevant entities. In tasks like video captioning, generic nouns (e.g., \textit{``video"}) serve as entities, enabling the entity-guided feature modulator to process the entire video (Figure \ref{fig:vidcap}). If no entities are present,  the modulator returns raw frame features to the short-term memory bank ensuring adaptability. For multiple non-distinguishing entities (Figure \textcolor{cvprblue}{12} in supplementary), the modulator processes all, while HierarQ prioritizes relevant relationships using the scene stream and long-term memory. 

\begin{figure*}
    \centering
    \includegraphics[width=.96\linewidth, height=4.9cm]{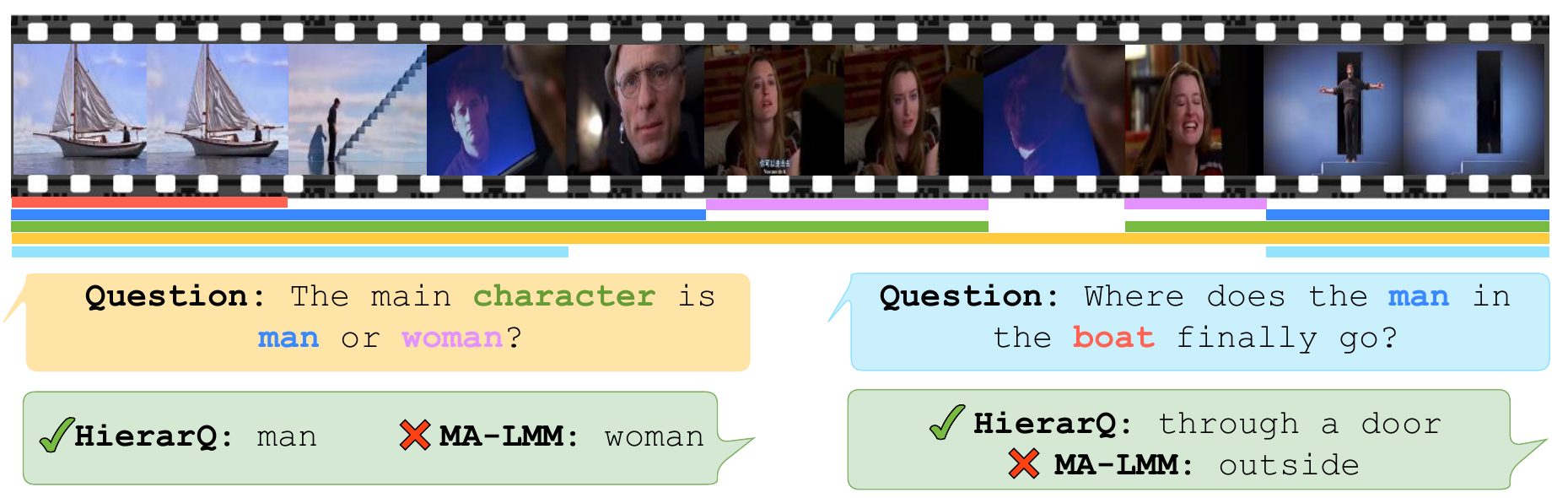}
    \includegraphics[width=.96\linewidth, height=4.6cm]{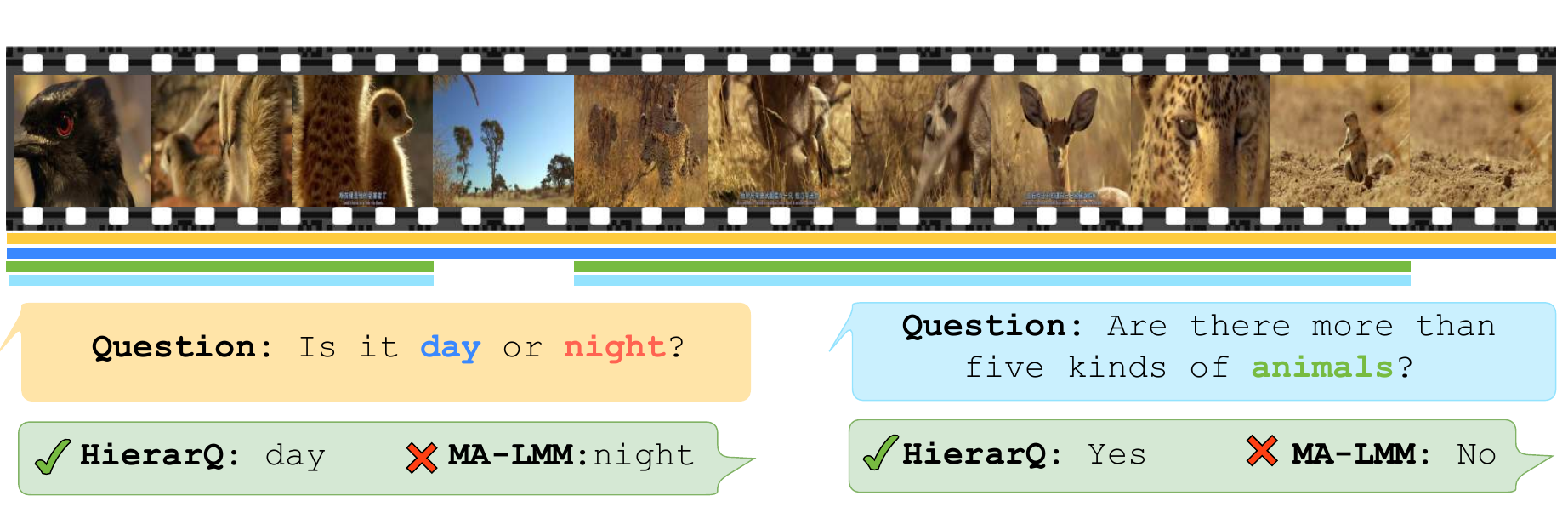}
    \includegraphics[width=.96\linewidth, height=4.6cm]{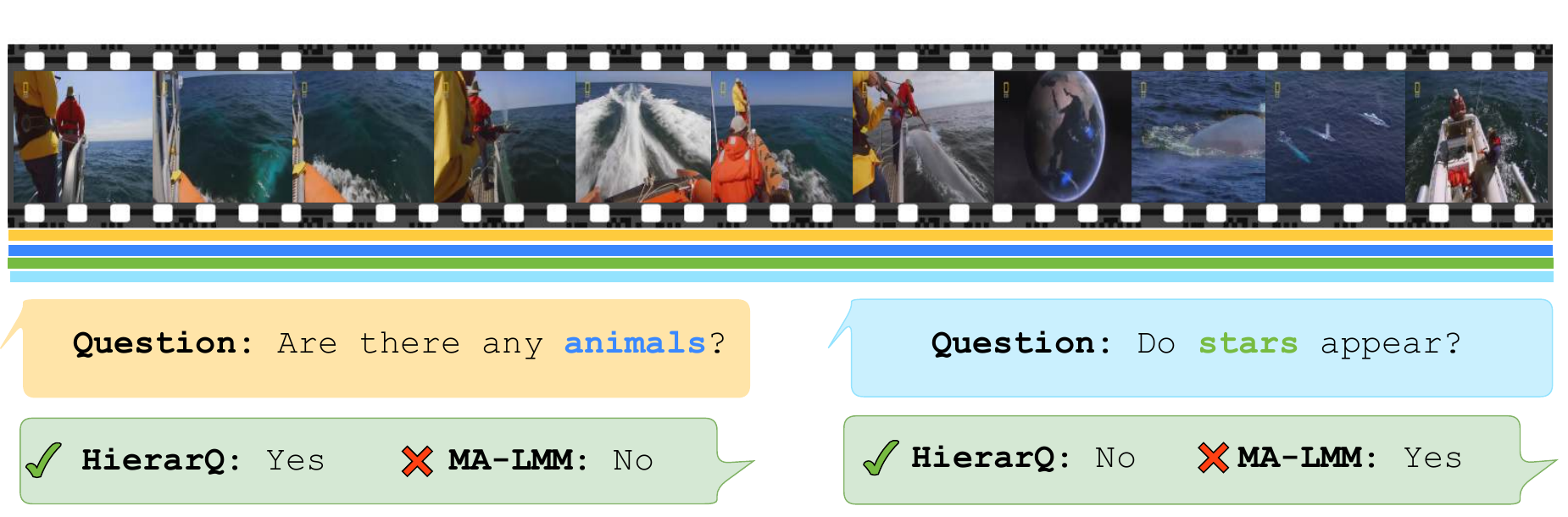}
    \caption{\textbf{Qualitative analysis of long-video question answering} on MovieChat-1k. Here, HierarQ adaptively focuses on task-relevant video segments, achieving a task-aware, comprehensive understanding. Color-coded frames are shown to demonstrate how entity-focused information complements the broader prompt-relevant context, enhancing overall video relevance and understanding.}
    \label{fig:qual-fig-sup}
\end{figure*}

\begin{figure*}[t!]
    \centering
    \includegraphics[width=.85\linewidth]{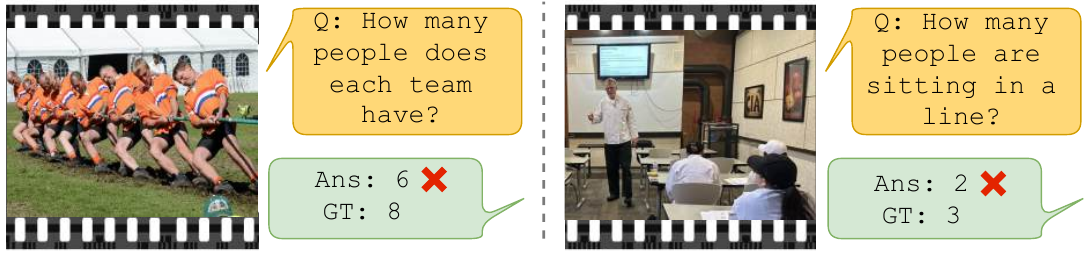}
    \caption{\textbf{Qualitative results of failure cases} on ActivityNet-QA.}
    \label{fig:failure}
\end{figure*}

\noindent
\textbf{Motivation for a two stream architecture.} The main motivation of the two-stream approach is to capture fine-grained entity details within a short context window separately from the broader scene-level understanding. The two streams complement each other (Table \ref{tab:ablation}) by preventing key-information loss and effective extended temporal relationship modeling. Since actions are inherently tied to entities, the entity stream captures not only the entities but also their interactions within its focus window, and the scene stream situates those interactions in a global context. While a verb-focused stream is an interesting idea, testing its inclusion resulted in similar performance ($0.53$\% drop on MSRVTT-QA) to the two-stream architecture. 

For medium to long-context understanding, key information may be scattered across time and at risk of being lost due to memory constraints. While the scene-stream is based on the full prompt and captures global context, it risks losing crucial short-term entity-level details over extended time. The entity-stream mitigates this by providing entity-specific information within a shorter temporal window as a complementary signal (Table \ref{tab:ablation}). Moreover, since it only focuses on entities in a short context, it is not over-crowded by other irrelevant information that might be present within that window. This balanced representation of local and global information enhances understanding. 

\section{Qualitative Analysis}
\label{sec:res_qual}
In Figure \ref{fig:qual-fig-sup}, we present a qualitative comparison of HierarQ and MA-LMM on the long-video question answering task using the MovieChat-1k dataset, which is the longest dataset among our benchmarks. The results highlight HierarQ's superior task-aware video understanding capabilities over MA-LMM. The two-stream task-aware feature modulator enables effective entity- and scene-level understanding through its dual-stream design with dedicated memory banks which further supports HierarQ to effectively model the temporal relationship between short and long-term contexts. 

For example, in the animal counting task, the entity-guided feature modulator along with the short-term memory bank helps track entity-specific details (e.g. ``\textit{animals}") across frames, while the prompt-guided feature modulator along with the long-term memory bank ensures continuity and accurate aggregation of historical information, enabling HierarQ to provide the correct answer. Similarly, the interplay between the entity- and scene-level Q-Formers inside the HierarQ allows nuanced reasoning, as seen in the ``\textit{man in the boat}" scenario, where HierarQ effectively models temporal relationships and historical context to deduce the correct outcome. 

In contrast, MA-LMM lacks task-awareness and treats all frames equally, relying on coarse memory compression that leads to errors in tasks requiring detailed and contextual understanding. Even in tasks requiring whole-video analysis, such as identifying the presence of animals or stars, HierarQ excels by leveraging its task-aware design and superior historical information retaining capability with the help of two level of memory banks and hierarchical Q-formers. HierarQ demonstrates a superior ability to understand the video by analyzing it holistically, effectively identifying semi-rare events such as the appearance of the whale, which occurs only a few times. In contrast, MA-LMM might be missing out that information due to coarse compression across longer timeline without task awareness. Additionally, HierarQ accurately detects the absence of stars, avoiding the potential bias in MA-LMM that associates the presence of Earth with stars due to its focus on the planet's view from space. By leveraging task-aware modulation, a hierarchical Q-Former, and memory integration, our framework dynamically models both short- and long-term temporal relationships, enabling a more accurate and comprehensive understanding of videos. 

On the contrary, Figure \ref{fig:failure} highlights HierarQ’s mispredictions due to ambiguous spatial arrangements.

\section{Additional Implementation Details}
\label{sec:imp_sup}
Table \ref{tab:arch-sup} outlines the architectural details of our framework. The hidden size of the feature modulators is aligned with the ViT's hidden size to ensure compatibility. Similarly, the additional attention submodules in the Scene-level Q-Former of HierarQ maintain the same number of attention heads and hidden size as the previous layers for consistency. 

\begin{figure*}
    \centering
    \includegraphics[width=\linewidth]{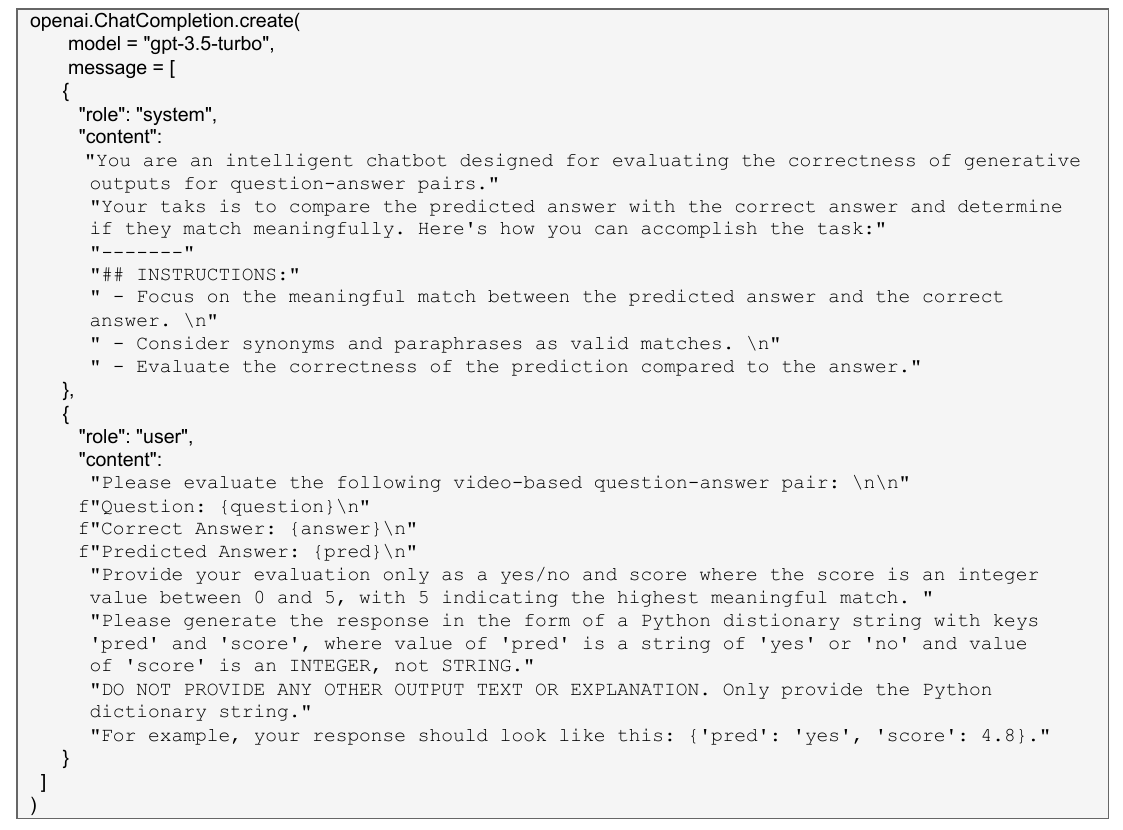}
    \caption{\textbf{Prompt for GPT 3.5 assisted evaluation} for the long-video question answering task.}
    \label{fig:gpt_eval}
\end{figure*}

Table \ref{tab:train-sup} provides the hyperparameter details of the training setup. Across all experiments, we employ cosine learning rate decay, and the frozen ViT and LLM components are converted to FP16 precision to optimize performance. For evaluation, we adhere to standard protocols across datasets following \cite{he2024ma}. We use $100$ frames as input for all datasets. The train-test split for all dataset is presented in Table \ref{tab:dataset}. One sample prompt for GPT-3.5-assisted evaluation is illustrated in Figure \ref{fig:gpt_eval}) which is used in long-video QA task evaluation.

\section{Future Work}
\label{sec:future}
Future work could focus on developing dynamic memory management strategies that prioritize frames and scenes based on task relevance. By introducing adaptive memory update mechanisms, it would be possible to selectively compress or discard less relevant information, optimizing memory usage while maintaining performance. For example, task-aware memory filters could assess the importance of incoming features and dynamically decide whether to store or discard them, allowing the model to concentrate on the most critical temporal or spatial details. To further improve scalability and reduce inference time, processing videos in smaller chunks and modeling inter-chunk relationships through advanced techniques such as hierarchical attention or transformer-based methods could be explored. These enhancements would aim to improve the efficiency and effectiveness of video analysis tasks across various domains.

\begin{table}[t!]
    \centering
    \small
    \caption{\textbf{Architectural details.}}
    \begin{tabular}{l|c|c}
        \hline
        \multirow{2}{*}{Hyper-parameters}&\multicolumn{2}{c}{\textbf{Task-aware Feature Modulator}}\\
         & Entity-guided & Prompt-guided\\
         \hline
         \# of layers & $2$ & $2$\\
         \# of attention heads & $8$ & $8$\\
         Hidden size & $1408$ & $1408$\\
         \hline
         \hline
         \multirow{2}{*}{Hyper-parameters}&\multicolumn{2}{c}{\textbf{HierarQ}}\\
         & Entity-level & Scene-level\\
         \hline
         \# of layers & $12$ & $12$\\
         \# of attention sub-modules & $2$ & $4$\\
         \# of attention heads & $12$ & $12$\\
         Hidden size & $768$ & $768$\\
         Cross attention frequency & $2$ &$2$\\
         \# of output query tokens & $32$ &$32$\\
         Memory bank length & $10$ & $10$ \\
         \hline

    \end{tabular}
    \label{tab:arch-sup}
\end{table}

\begin{table}[t!]
    \centering
    \caption{\textbf{Hyper-parameters for training.}}
    \begin{tabular}{l|c}
        \hline
         Hyper-parameters& Value \\
        \hline
        Patch size &$14 \times 14$\\
        Frame resolution &$224 \times 224$\\
        Training epoch &$20$\\
        Batch size &$32$\\
        Learning rate &$1e$-$5$\\
        Weight decay &$0.05$\\
        AdamW $\beta$ &$[0.9, 0.999]$\\
        LoRA rank &$32$\\
        Beam size & $5$\\
        \hline
    \end{tabular}
    \label{tab:train-sup}
\end{table}
\begin{table}[t]
    \centering
    \caption{\textbf{Dataset statistics.} Here QA pairs denote question-answer pair only applicable for video question answering task.}
    \begin{tabular}{c|c|c|c}
        \hline
         Dataset & Split & \# Videos & \# QA pair  \\
         \hline
         \multicolumn{4}{c}{\textbf{Task: Video Understanding}}\\
         \hline
         \multirow{3}{*}{LVU} & train & $6927$& -\\ 
         & validation & $1477$&-\\ 
         & test & $1394$&-\\
         \hline
         \multirow{2}{*}{Breakfast} & train & $8451$&-\\ 
         & test & $2816$&-\\
         \hline
         \multirow{2}{*}{COIN} & train & $9030$ & -\\ 
         & test & $2797$ & -\\
         \hline
         \hline
         \multicolumn{4}{c}{\textbf{Task: Video Question Answering}}\\
         \hline
         \multirow{3}{*}{MSRVTT-QA} & train & $6513$ & $158581$\\ 
         & validation & $2990$ & $12278$\\ 
         & test & $497$ & $72821$\\
         \hline
         \multirow{3}{*}{MSVD-QA} & train & $1200$& $30933$\\ 
         & validation & $250$ & $6415$ \\ 
         & test & $520$ & $13157$ \\
         \hline
         \multirow{3}{*}{ActivityNet-QA} & train & $3200$ & $32000$\\ 
         & validation & $1800$ & $18000$\\ 
         & test & $800$ & $8000$\\
         \hline
         \multirow{3}{*}{MovieChat-1k} & train & $800$ & $10400$\\ 
         & validation & $100$ & $1300$\\ 
         & test & $100$& $1300$\\
         \hline
         \hline
         \multicolumn{4}{c}{\textbf{Task: Video Captioning}}\\
         \hline
         \multirow{3}{*}{MSRVTT} & train & $6513$ &-\\ 
         & validation & $2990$ &- \\ 
         & test & $497$  & -\\
         \hline
         \multirow{3}{*}{MSVD} & train & $1200$ &-\\ 
         & validation & $250$ &-\\ 
         & test & $520$ &- \\
         \hline
         \multirow{3}{*}{YouCook2} & train & $1333$ &-\\ 
         & validation & $457$ &-\\ 
         & test & $210$ &-\\
         \hline
    \end{tabular}

    \label{tab:dataset}
\end{table}